\pdfoutput=1
\documentclass[letterpaper, 10 pt, journal, twoside]{IEEEtran}
\IEEEoverridecommandlockouts
\usepackage{amsmath,amssymb, amsfonts, bm} % for \boldsymbol macro
\usepackage{url}
\usepackage{doi}
\usepackage{newclude}
\usepackage{graphicx}
\usepackage{tabularx}
\usepackage[font={small}]{caption}
\usepackage{subcaption}
\usepackage{dblfloatfix}
\usepackage[linesnumbered,ruled,vlined]{algorithm2e}
\usepackage{multirow}
\usepackage{color}
\usepackage{cite}
\usepackage{hyperref}

\begin{document}
% The paper headers
% \markboth{Journal name and date}%
% {Shell \MakeLowercase{\textit{et al.}}: }
\title{Learning Problem Decomposition for Efficient Sequential Multi-object Manipulation Planning}
\author{Yan Zhang, Teng Xue, Amirreza Razmjoo, Sylvain Calinon%
\thanks{Manuscript received: August 13, 2025; Accepted October 16, 2025. This paper was recommended for publication by Editor Aleksandra Faust upon evaluation of the Associate Editor and Reviewers’ comments.}
\thanks{The authors are with the Idiap Research Institute, Switzerland and also with the Ecole Polytechnique F{\'e}d{\'e}rale de Lausanne (EPFL), Switzerland; Email: firstname.lastname@idiap.ch}%
\thanks{This work was supported by the State Secretariat for Education, Research and Innovation in Switzerland for participation in the European Commission’s Horizon Europe Program through the INTELLIMAN project (\url{https://intelliman-project.eu/}, HORIZON-CL4-Digital-Emerging Grant 101070136) and the SESTOSENSO project (\url{http://sestosenso.eu/}, HORIZON-CL4-Digital-Emerging Grant 101070310). We also acknowledge support from the China Scholarship Council (Grant No. 202106230104).}%

\thanks{Digital Object Identifier (DOI): see top of this page.}
}    

\markboth{IEEE Robotics and Automation Letters. Preprint Version. Accepted October, 2025}
{ZHANG \MakeLowercase{\textit{et al.}}: Learn2Decompose}

% make the title area
\maketitle

\begin{abstract}
    We present an efficient task and motion replanning approach for sequential multi-object manipulation in dynamic environments. Conventional Task And Motion Planning (TAMP) solvers experience an exponential increase in planning time as the planning horizon and number of objects grow, limiting their applicability in real-world scenarios. To address this, we propose learning problem decompositions from demonstrations to accelerate TAMP solvers. Our approach consists of three key components: \textit{goal decomposition learning}, \textit{computational distance learning}, and \textit{object reduction}. Goal decomposition identifies the \textit{necessary} sequences of states that the system must pass through before reaching the final goal, treating them as subgoal sequences. Computational distance learning predicts the computational complexity between two states, enabling the system to identify the temporally closest subgoal from a disturbed state. Object reduction minimizes the set of active objects considered during replanning, further improving efficiency. We evaluate our approach on three benchmarks, demonstrating its effectiveness in improving replanning efficiency for sequential multi-object manipulation tasks in dynamic environments. Accompanying website: \href{https://sites.google.com/view/learn2decompose?usp=sharing}{https://sites.google.com/view/learn2decompse}
\end{abstract}


\renewcommand{\textcolor}[2]{#2}

\IEEEpeerreviewmaketitle

\section{Introduction}
\IEEEPARstart{N}{umerous} daily activities, from cooking \cite{yang2023sequence} to furniture assembly \cite{suarez2018can}, require sequential manipulation planning over multiple objects. These tasks often involve planning over various interactions among robots and objects under spatial-temporal constraints, and over multiple steps with sparse rewards \cite{hartmann2022long, du2023guiding,di2023towards}, posing significant challenges for autonomous robotic solutions. Task and Motion Planning (TAMP) has emerged as a powerful framework for tackling complex sequential multi-object manipulation tasks \cite{yang2023sequence,hartmann2022long}. TAMP integrates task planning with motion planning by combinatorially searching sequences of abstracted actions and corresponding motion parameters \cite{garrett2021integrated, toussaint2015logic, garrett2020pddlstream}. Consequently, as the planning horizon and the number of environmental objects grow, TAMP 
\begin{figure}[thb]
    \centering
    \begin{subfigure}{0.32\linewidth}
        \centering
        \includegraphics[width=\linewidth]{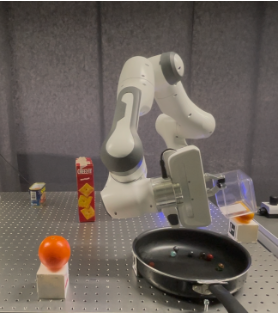}
        \caption{}
        \label{fig:real_kitchen_c}
    \end{subfigure}
    \begin{subfigure}{0.32\linewidth}
        \centering
        \includegraphics[width=\linewidth]{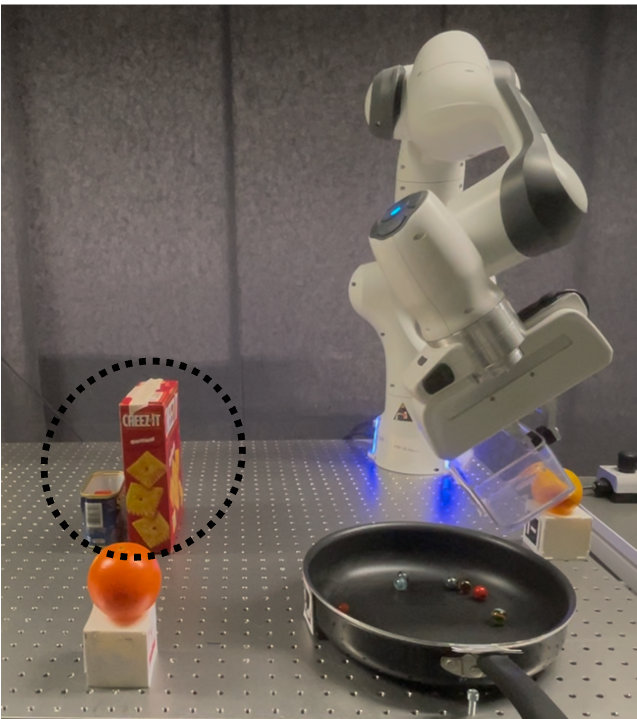}
        \caption{}
        \label{fig:real_kitchen_d}
    \end{subfigure}
    \begin{subfigure}{0.32\linewidth}
        \centering
        \includegraphics[width=\linewidth]{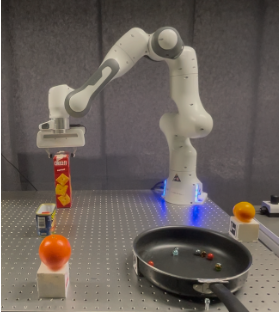}
        \caption{}
        \label{fig:real_kitchen_e}
    \end{subfigure}
    \caption{The Franka Emika robot arm responds robustly to disturbances (black dashed line) in the positions of the crackers box and meat can, using our efficient task and motion replanning approach, which is enabled by integrating learned problem decomposition from demonstrations.}
    \label{fig:reactive_tamp_franka_intro}
    \vspace{-0.8cm}
\end{figure}
solvers require exponentially longer planning time \cite{yang2023sequence, silver2021planning, Kaelbling2011hierarchical}. This time-intensive planning thus poses challenges when using TAMP solvers in closed-loop systems requiring to efficiently react to disturbances in real-world scenarios. 

A considerable number of manipulation tasks typically require passing through several subgoals in sequence to achieve the final task goal. For example, cooking a meal often involves preparing various ingredients in a specific order as dictated by a recipe (see Figure  \ref{fig:reactive_tamp_franka_intro}). On one hand, finding solutions that follow these subgoal sequences in the correct order requires identifying a small subset of solutions from a vast solution space, which significantly increases planning time. On the other hand, we argue that these subgoal sequences can be extracted from demonstrations and used to decompose complex tasks into simpler subtasks. For instance, determining the correct order for processing ingredients without a recipe would involve extensive trial-and-error planning. However, based on prior experience preparing the same meal, one can summarize a recipe and focus only on reaching the next closest subgoal in that sequence—thereby simplifying the planning process. More importantly, these summarized subgoal sequences exhibit strong robustness to variations in state and environment. For example, a meal can typically be prepared by following the same recipe, regardless of the initial placement of ingredients, the layout of the kitchen, or the shapes of objects. This intuition supports our argument that learning subgoal sequences not only enables goal decomposition but also imparts robustness to task variations. We further argue that this resulting decomposition and robustness can greatly facilitate the design of efficient TAMP solvers for sequential multi-object manipulation planning under disturbances.

We proposed to enhance classical TAMP solvers with problem decompositions learned from demonstrations. The proposed problem decomposition consists of three key components: \textit{goal decomposition}, which analyzes the collected demonstrations to identify the \textit{necessary} sequences of states the system must pass through before reaching the final goal, treating them as subgoal sequences; \textit{\textcolor{blue}{computational} distance learning}, which uses the same set of demonstrations to train a Graph Neural Network (GNN) that predicts the importance of environmental objects and uses the number of important objects as a \textcolor{blue}{computational} distance metric to select the closest subgoal from a disturbed state; and \textit{object reduction}, which runs multiple TAMP solvers with different object sets to find a feasible plan that connects the disturbed state to the predicted closest subgoal, improving replanning efficiency.

In summary, the main contributions of this paper are: 1) We propose a novel subgoal formulation and generation approach, where partial state sequences that must be sequentially traversed are considered as subgoal sequences for goal decomposition, and we generate them from demonstrations by casting this as a sequential pattern mining problem; 2) We introduce a computational distance learning method that approximates computational distance between states via object importance prediction, facilitating the selection of the closest subgoal for efficient task and motion replanning; 3) We develop an object reduction mechanism that preserves solution completeness while accelerating replanning efficiency; 4) Integrating above three learned problem decomposition modules with classical TAMP solvers, we introduce an efficient task and motion replanning framework for sequential multi-object manipulation planning in dynamic environments.

\section{Related Work}
\subsection{Learning Subgoals for Long-horizon Planning}
Our motivation of learning subgoal sequences from demonstrations is based on previous research works that indicate subgoals can be used to decompose a long-horizon planning problem into simpler subproblems, thus accelerating long-horizon planning efficiency \cite{Xu2018NTP, xu2019regression}. However, these works assume that subgoals are included in the training dataset and train a neural network to predict online the next subgoals for accelerating task planners. Similarly, with internet-level dataset, Large Language Models (LLMs) can be used to infer subgoals to accelerate the learning efficiency for solving complex sequential manipulation tasks \cite{di2023towards, du2023guiding}. Our work is complementary to such works in the sense that we assume no access to such subgoals in the dataset and aim to generate such subgoals from unlabeled dataset in an unsupervised way. Recently, \textit{Elimelech et al.}\cite{Elimelech2023extracting} identifies critical states from past experiences and use segmented state trajectories as higher-level action abstractions for speeding up task planning \cite{ Elimelech24icra}. Our approach mainly differs from \cite{Elimelech2023extracting,Elimelech24icra} in the definition and usage of subgoals. In their methods, subgoals are defined as critical states that maximize abstraction level and minimize state trace segmentation. The traces between consecutive subgoals are then used as higher-level symbolic actions to decompose tasks. In contrast, we identify \textit{partial} states that must be traversed to reach the task goal as subgoals, and use them as checkpoints during replanning—without reusing the state traces between them—as such traces may need to be replanned in integrated TAMP. \textit{Levit et al.}~\cite{levit2024solving} suggested identifying subgoals in the bottleneck region of the environment to improve the resolution of 2D puzzle problems with optimization-based TAMP solvers. In their work, subgoals are generated based on environment geometry to alleviate local optima issues in optimization-based motion planning. Differently, we define must-pass states as subgoals and use them to shorten the planning horizon and enhance the planning efficiency of TAMP solvers.

\subsection{Task and Motion Planning}
Our work falls within the domain of learning for Task and Motion Planning (TAMP), where machine learning methods are applied to leverage previous experiences to improve the planning efficiency of TAMP solvers. Existing works include learning value functions as cost-to-go heuristics \cite{kim2020learning}, learning action feasibility \cite{driess2020deep, park2022scalable, yang2023sequence} to prune infeasible and computationally expensive motion-planning validations, and identifying important objects \cite{silver2021planning} or their associated streams \cite{khodeir2023learning}, all of which significantly speed up TAMP solvers. Our approach differs from these methods by proposing to learn necessary subgoal sequences as guidance and a computational distance metric for selecting the temporally closest subgoal from a disturbed state, thereby minimizing the planning horizon and number of objects for TAMP solvers, thus enhancing their planning efficiency.

To achieve fast replanning, \textit{Migimatsu et al.}~\cite{migimatsu2020object} introduced an object-centric TAMP approach, where motion trajectories are executed in the Cartesian coordinates of the corresponding objects, demonstrating reactivity to motion-level disturbances. To address both task- and motion-level disturbances, \cite{harris2022fc, zhang2024multi} propose an offline construction of a set of prioritized action plans, which can then be dynamically switched online for fast replanning. The success of this pipeline relies on the completeness of the constructed plan library and cannot effectively handle disturbed states that fall outside the predefined plans. \textit{Xue et al.}~\cite{xue2024d} explore goal backpropagation to improve the planning efficiency of TAMP solvers and formulate their approach in a closed loop to handle various disturbances. Their method relies on the effectiveness of the back-propagation heuristic, which can be difficult to generalize across different domains. 

In terms of accelerating TAMP solvers via planning horizon reduction, \textit{Castaman et al.}~\cite{castaman2021receding, braun2022rhh} propose Receding-Horizon TAMP (RH-TAMP) approaches that iteratively solve a reduced problem over a fixed receding horizon. However, fixed receding horizons often fail to capture the full-horizon spatial-temporal constraints, which can result in increased infeasibility of action skeletons. The work most closely related to ours is Logic Learning from Demonstrations (LogicLfD) \cite{zhang2024logic}, which reduces the planning problem to a partial problem by running a TAMP solver to reach the demonstrated trajectory as fast as possible. In contrast, the method presented in this paper represents demonstrations as sequences of necessary subgoals, which enables to accelerate TAMP solvers by reducing both the planning horizon and the number of objects considered. This results in superior reactivity in manipulation tasks with many objects and long planning horizons.

\begin{figure}[thb!]
    \centering
    \vspace{-0.3cm}
    \includegraphics[width=\linewidth]{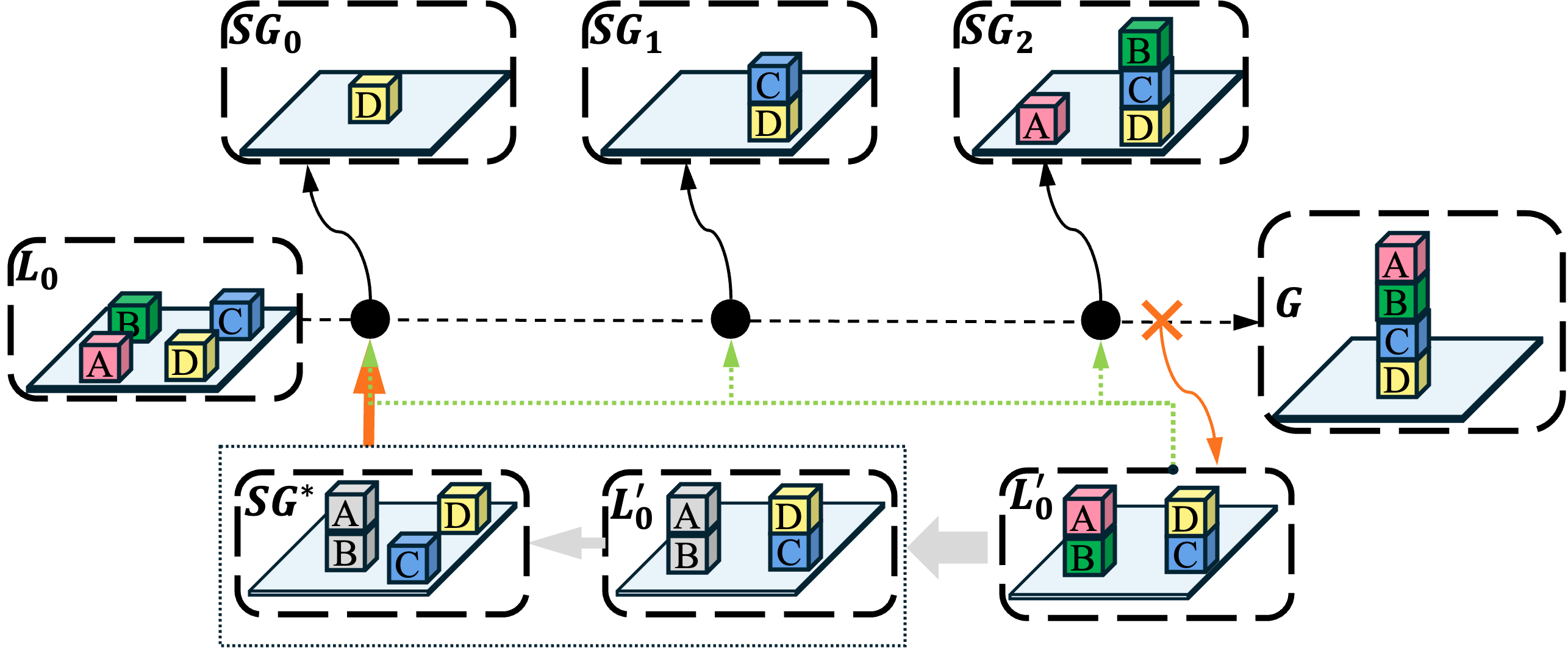}
    \caption{An overview of our efficient planner, which integrates \textit{goal decomposition}, \textit{computational distance}, and \textit{object reduction} with classical TAMP solvers to enable efficient replanning under online disturbances (orange 'x' symbol). Goal decomposition identifies the \textit{necessary} subgoal sequences $(\bm{SG_{0}} \rightarrow \bm{SG_{1} \rightarrow \bm{SG_{2}}})$ offline from demonstrations. The computational distance metric is trained offline using the same set of demonstrations and is used to measure the distance (green arrows) from a disturbed state $\bm{L_{0}^{'}}$ to the subgoals (black circles), enabling the identification of the closest subgoal $\bm{SG^{*}}=\bm{SG_{0}}$. Object reduction (area with dashed line) considered grey cubes as fixed obstacles while replanning from $\bm{L_{0}^{'}}$ to $\bm{SG^{*}}$ so that only necessary objects are involved in. The whole process is repeated iteratively until the final task goal is achieved. }
    \label{fig:pipeline}
    \vspace{-0.5cm}
\end{figure}

\section{Method}
Figure~\ref{fig:pipeline} presents an overview of our efficient task and motion replanning approach, which integrates problem decomposition with TAMP solvers (e.g., PDDLStream~\cite{garrett2020pddlstream}, used in this paper) for efficient sequential multi-object manipulation planning. Our problem decomposition framework consists of three components: \textit{goal decomposition learning}, \textit{computational distance learning}, and \textit{object reduction}. We describe each of these components in detail below. As an extension of Figure~\ref{fig:pipeline}, we provide a pseudocode description of our efficient task and motion replanning approach in Appendix~\ref{app:pseudocode} to offer a more detailed explanation of our algorithms.

\subsection{Demonstration Generation}
\begin{figure}[t!hb]
    \centering
    % \vspace{-0.3cm}
    \includegraphics[width=0.8\linewidth]{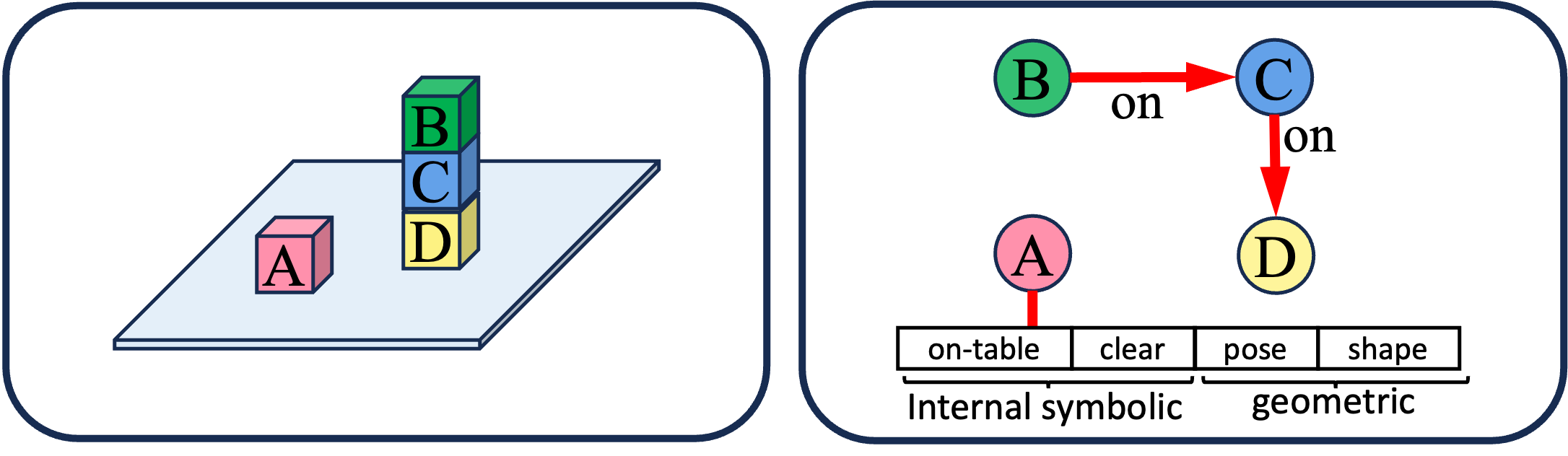}
    \caption{Illustration of objects' configurations at specific time step and its graph representation.}
    \label{fig:scene_graph}
    \vspace{-0.5cm}
\end{figure}

The left image in Figure~\ref{fig:scene_graph} shows the configuration of objects at a specific time step $t$. Boolean symbolic variables $s_t \in \mathcal{S}$ are used to define the symbolic state $\bm{s}_t$ of the environment at time step $t$. These symbolic variables capture internal properties, geometric information, and spatial relationships between objects—for example, \texttt{(clear B)}, \texttt{(atPose B p1)}, and \texttt{(on B C)} describe whether object \texttt{B} is clear, its pose in the environment, and whether it is on top of another object \texttt{C}, respectively. A symbolic state $\bm{s}_t$ is an itemset of Boolean symbolic variables drawn from the set $\mathcal{S}$, representing symbolic description of all environment objects at time step $t$.

We assume access to a set of demonstrations that successfully achieve the task goal $\bm{G}$ across $N$ distinct initial states. In this paper, we use PDDLStream~\cite{garrett2020pddlstream} to collect $N=40$ demonstrations for all tasks in Section~\ref{sec:experiments}. The different initial states are generated by randomizing the logical states of environmental objects and their geometric poses. However, these demonstrations could also be obtained from human demonstrations, internet datasets, or past planning experiences of any planner, as long as they include the geometric states of 
environment objects (e.g., poses, shapes) so that they can be translated into symbolic states using a predefined state abstraction. We represent the unified dataset of demonstrations as $\mathcal{P} = \{\mathcal{P}^{s}, \mathcal{P}^{g}\}$, where $\mathcal{P}^{s}$ and $\mathcal{P}^{g}$ denote symbolic and geometric state trajectories, respectively. The $i$th symbolic trajectory $\mathcal{P}_i^s$ is a sequence of symbolic states $\bm{s}_{t=1,\dots,k_i}$ of length $k_i$, while the corresponding geometric trajectory $\mathcal{P}_i^g$ captures the pose trajectories of all objects over time.

\subsection{Goal Decomposition Learning}
With the demonstrations, our goal is to identify sequences of \textit{partial states} in $\mathcal{P}^{s}$ that are consistently traversed to reach the target task goal—regardless of the initial state or the specific action plan chosen. Here, \textit{partial states} refer to symbolic states of a subset of environmental objects in the planning domain. For instance, in $\bm{SG_0}$ of Figure~\ref{fig:pipeline}, only \texttt{(ontable, D)} and \texttt{(clear, D)} are considered as \textit{partial states}. \textcolor{blue}{Although our symbolic states also include geometrical information with geometrical predicates, these symbolic states related to cube D, such as \texttt{(atpose, D, p1)}, and those indicating relationship between cube D and other cubes are excluded from subgoal $\bm{SG_0}$ because they are not essential for achieving the final goal $\bm{G}$.} Since these partial states must be visited in a specific order to reach the goal, we expect to observe the same sequences of partial states across all demonstrations. We therefore formulate the goal decomposition problem as follows: given a set of demonstrations, identify sequences of partial states that are consistently observed across all demonstrations as subgoal sequences.

Given that the symbolic state $\bm{s}_t$ is represented as an itemset of Boolean symbolic variables $s$, the symbolic trajectory $\mathcal{P}^s$ can be expressed as $N$ sequences of itemsets $\mathcal{H} = \{\bm{s}_{t=1,\dots,k_i}\}_{i=1,\dots, N}$. We then identify the sequences of itemsets that are common across all symbolic trajectories by formulating it as a sequential pattern mining (SPM) problem \cite{fournier2017survey}. In this paper, we adopt PrefixSpan \cite{pei2004mining}, a widely used SPM algorithm known for its simplicity and flexibility in incorporating constraints \cite{fournier2017survey}. For more details on the algorithm, refer to \cite{pei2004mining, fournier2017survey}.

PrefixSpan operates on the sequence database $\mathcal{H}$ by recursively projecting only those parts of the database that are relevant to a given prefix pattern. This thus greatly reduces the search space and efficiently identifies all the common sequences of itemsets $\{\tilde{\bm{s}}_{\bm{t}_{m}}\}_{m=1, M}$ that meet or exceed a minimum occurrence frequency ratio, defined as \textit{min-support}.  Here, $\bm{t}_m$ is a list of time index, and each $\tilde{\bm{s}}_{\bm{t}_m}$ denotes a sequence \textit{partial states} of symbolic states $\bm{s}_{\bm{t}_m}$ at key time indices in $\bm{t}_m$. $M$ is the total number of discovered sequential patterns. We set the minimum support threshold (\textit{min-support}) to 0.9, targeting those sequential itemsets that appear in at least 90\% of the demonstrations.

\begin{figure}[t!]
    \centering
    \includegraphics[width=\linewidth]{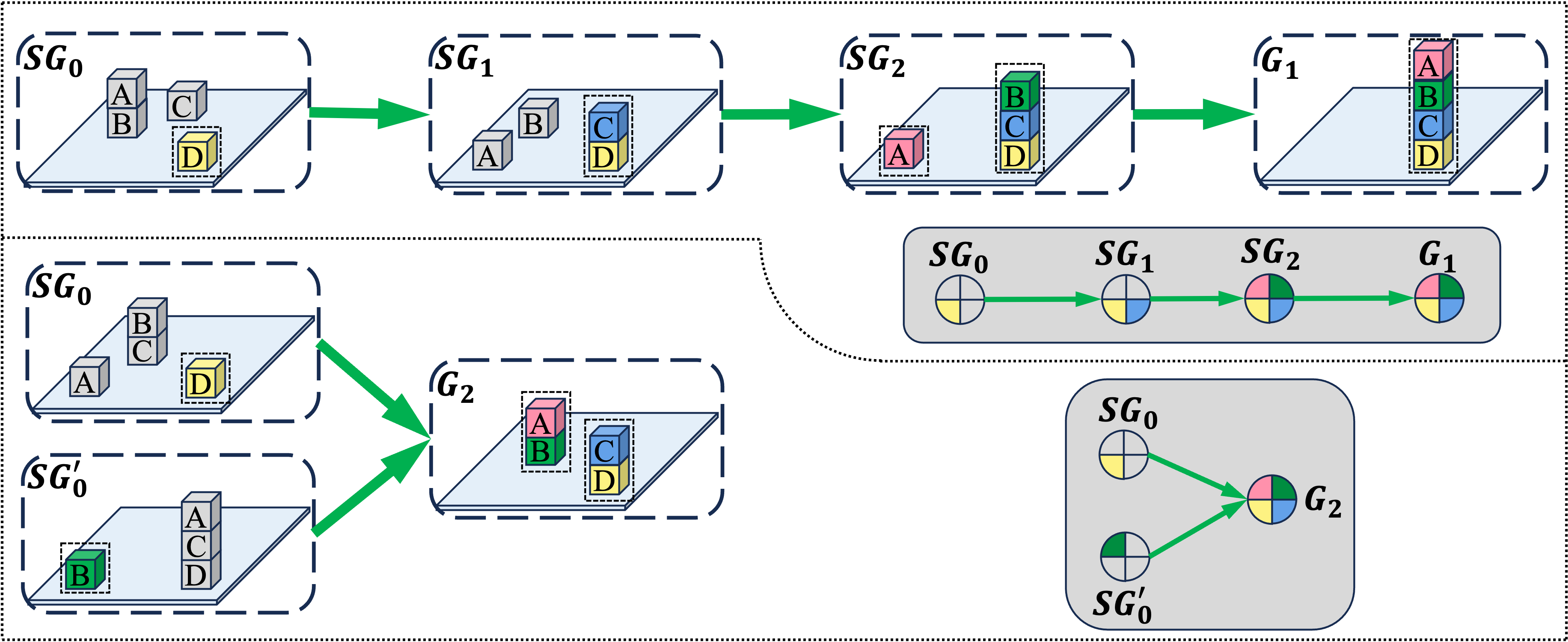}
    \caption{Illustration of subgoal sequences for two task goals $\bm{G}_1, \bm{G}_2$ in tower construction toy example.}
    \label{fig:goal_decomposition}
    \vspace{-0.5cm}
\end{figure}

Using PrefixSpan, we generate a set of itemset sequences including both the target subgoal sequences $\bm{SG} = \{\bm{SG}_0,\dots, \bm{SG}_Z\}$ and all its subsequences, where $Z$ is an unknown variable representing the length of the target sequence. We iteratively compare the generated sequential patterns to filter out all the subsequences. Correspondingly, we will generate one or multiple sequences of subgoals based on the decomposability of the task goal. Figure \ref{fig:goal_decomposition} illustrated subgoal sequences of two different task goals. The top sequence shows the subgoal sequence for task goal $\bm{G}_1$ where blocks should be stacked in a linear sequential order, from $\bm{SG}_0$ to $\bm{G}_1$. In this sequence, each subgoal (partial symbolic states of colorful blocks within dashed line) captures a partial state of the environment, with only the colored blocks being relevant for the subgoal, while grey blocks can be at any configuration. Areas with grey background show the corresponding Directed Acyclic Graph (DAG) representation of the sequential structure of the subgoals for $\bm{G}_1$; In contrast, the lower sequence for $\bm{G}_2$ illustrates a multi-modal sequential solution. We can only \texttt{stack} \texttt{A/C} \texttt{on} \texttt{B/D} once \texttt{B/D} are \texttt{clear}; However, to achieve $\bm{G}_2$, it does not matter whether \texttt{D} becomes \texttt{clear} before \texttt{B} becomes \texttt{clear}, or even after \texttt{A} is stacked on \texttt{B}, as the only constraint is that \texttt{D} must be \texttt{clear} before stacking \texttt{C} on it. Consequently, we will observe state trajectories where \texttt{D} becomes \texttt{clear} before or after \texttt{B} becomes \texttt{clear} in demonstrations. PrefixSpan thus identifies that different subgoal sequences ($\bm{SG}_0$ and $\bm{SG}_{0}^{'}$) can lead to the same final goal $\bm{G}_2$. This shows that in the same planning domain, different task goals have different numbers of subgoal sequences.

In Appendix~\ref{app:goal_decomposition}, we provided formal analysis on why and when task goals are decomposable, aiming at providing more details about the motivation and formulation of our goal decomposition method.

\subsection{Computational Distance Learning}
\subsubsection{Graph-based Demonstration Representation}
We represent the symbolic and geometric states of demonstrations $\mathcal{P}$ at each time step using a graph structure. In this graph, each object is represented as a node. Symbolic states that describe relationships between objects (e.g., \texttt{(onblock, B, C)}) are encoded as edge features. Other symbolic states that describe intrinsic properties of individual objects (such as \texttt{(clear, A)}), along with geometric information (e.g., pose, shape), are incorporated as node features. The right image in Figure~\ref{fig:scene_graph} show an example of the constructed graph structure.

\subsubsection{GNN-based Computational Distance Learning} Computational distance refers to the computational time required to find a feasible solution from a disturbed state to a subgoal or the task goal \cite{Myers2024learning}. The exact computational time cost depends on several factors, such as the configurations of the environment at the initial and goal states, the state and action abstraction in TAMP solvers, the corresponding streams \cite{garrett2020pddlstream} or motion planners, and hardware. These factors make it infeasible to design a general computational distance function that predicts computational time precisely across different problems. Therefore, we propose to approximate the relative computational distance\footnote{We define the actual system state which is different from the expected state after executing the target action as a disturbed state. The system starts to conduct task and motion replanning once detected a disturbed state.} $d(\bm{s}_1, \bm{SG})$ with the number of the important objects required for transitioning from the disturbed state $\bm{s}_1$ to a subgoal $\bm{SG}$. The motivation behind this representation is that planning time typically increases exponentially with the number of objects involved and the overall planning horizon. Generally, the more objects are involved, the longer the planning sequence is, or the more motion parameters must be determined during the motion planning phase. Thus, we posit that the number of important objects can serve as a useful indicator of relative \textcolor{blue}{computational} distance.

To learn this relative \textcolor{blue}{computational} distance metric, we segment demonstrations $\mathcal{P}$ based on the generated subgoal sequences $\bm{SG}$. Each segment thus describes the transition process from one subgoal $\bm{SG}_i$ to the next or a later subgoal $\bm{SG}_{i+1}$ or $\bm{SG}_{i+z}$, where $z \in [1, Z-i]$. In each segment, we identify important objects $\mathcal{O}^{*}$ for each subgoal transition process by analyzing changes in object states: objects whose states change during the transition are considered important. We then construct a graph-based dataset, where each input consists of the graph representations of the initial and goal states of a segment. The output is an importance score for each object: objects in $\mathcal{O}^{*}$ are assigned a score of 1, while others receive a score of 0.

We train a GNN model on this dataset to predict object importance. During online execution, when a disturbance is detected, we compute the \textcolor{blue}{computational} distance $d(\bm{s}_1, \bm{SG}_i)$ from the current disturbed state $\bm{s}_1$ to each subgoal $\bm{SG}_i$ based on the number of predicted important objects (i.e., those with scores greater than 0.9). The subgoal with the minimal predicted distance and earliest position in the subgoal sequence is considered the temporally closest subgoal $\bm{SG}^{*}$.

\begin{figure*}[t!hb]
    \centering
    \begin{subfigure}{0.30\linewidth}  % Adjusted width for better visibility
        \centering
        \includegraphics[width=\linewidth]{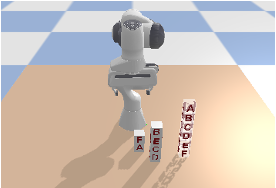}
        \caption{Tower construction}
        \label{fig:exp1}
    \end{subfigure}%
    \hspace{0.02\linewidth}
    \begin{subfigure}{0.30\linewidth}
        \centering
        \includegraphics[width=\linewidth]{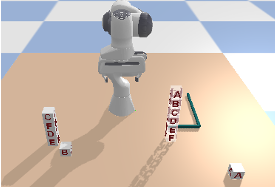}
        \caption{Tool use}
        \label{fig:exp2}
    \end{subfigure}%
    \hspace{0.02\linewidth}
    \begin{subfigure}{0.30\linewidth}
        \centering
        \includegraphics[width=\linewidth]{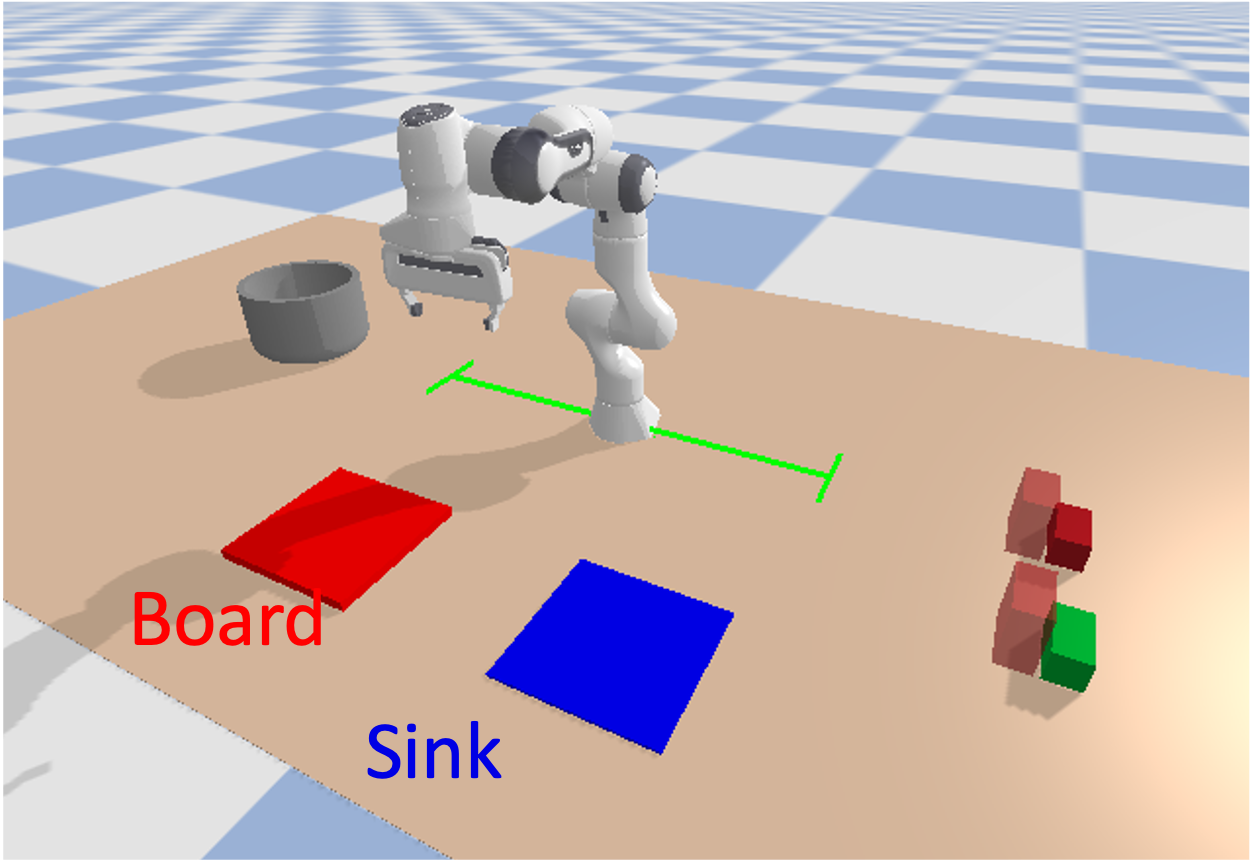}
        \caption{Cooking}
        \label{fig:exp3}
    \end{subfigure}
    \hspace{0.02\linewidth}
    \caption{In (a) and (b), transparent blocks indicate a representative task goal. In the Cooking domain, the Franka manipulator is equipped with a virtual mobile base, enabling movement along the Y-axis within the boundary indicated by green lines.}
    \label{fig:experimental_setups}
    \vspace{-0.5cm}
\end{figure*}

This relational representation is similar to the one presented by \textit{Silver et al.}\cite{silver2021planning} for predicating the importance of environmental objects in task planning. However, our approach differs in two key ways: (1) we incorporate the geometry properties of objects also as part of the node properties so that the trained GNN model aligns with TAMP solvers; (2) \textcolor{blue}{we employ the GNN model not only to approximate the corresponding computational distance but also to predict important objects for object reduction later}. Correspondingly, the advantages of our approach to formulate \textcolor{blue}{computational} distance are twofold. First, by using the same trained GNN model, we can simultaneously predict both the \textcolor{blue}{computational} distance for choosing the closest subgoals and use the predicted important objects for \textit{object reduction} while replanning to the predicted closest subgoals. Second, since we prioritize reaching the closest subgoal with the fewest important objects, the prediction accuracy for subsequent subgoals—those requiring prediction over longer horizons—becomes less critical. This results in a more simplified graph machine learning problem and alleviate the issue when requiring GNN to capture global long-horizon dependency among nodes \cite{di2023over}.

\subsection{Object Reduction}
In this section, we introduce an object reduction strategy to augment TAMP solvers, aimed at improving replanning efficiency when replanning to reach the predicted closest subgoal $\bm{SG}^{*}$. Traditionally, a TAMP solver seeks a feasible plan $\bm{\tau}$ for the problem $P(\bm{s}_1, \bm{SG}^{}, \mathcal{O})$, where the planner considers the full set of environmental objects $\mathcal{O}$ to connect the disturbed state $\bm{s}_1$ to the target subgoal $\bm{SG}^{}$. However, many objects in the environment may be irrelevant to achieving the subgoal, and reasoning over all of them introduces unnecessary computational overhead.

To enable object reduction, we leverage the predicted importance scores from the trained GNN model to dynamically construct reduced object sets for replanning. We begin with an initial importance threshold $\epsilon = 0.9$ and generate a descending sequence of thresholds $\{\epsilon, \epsilon^2, \dots, \min(\epsilon^i, 0.01), 0\}$ with 6 elements. For each threshold, we select a subset of objects $\mathcal{O}_i^{*}$ whose predicted importance scores exceed the threshold, and treat all remaining objects as fixed obstacles. This results in a set of subproblems $\{P_i\}_{i=1,\dots,6}$ with progressively larger object sets $\{\mathcal{O}_{i}^{*}\}_{i=1,\dots,6}$ but identical initial and goal states $\bm{s}_1$, $\bm{SG}^{*}$. We perform parallel replanning for each subproblem. If any of the subproblems returns a feasible plan, all other processes are terminated. This ensures that we prioritize solving the most efficient, object-reduced problem first, while still preserving the ability to solve the full problem if necessary.

Incorrect predictions from the GNN may exclude critical objects, potentially rendering a subgoal unreachable even if it is otherwise feasible. Moreover, in motion planning, infeasibility due to missing objects can be difficult to detect and may lead to unbounded planning time. To mitigate this, we include a minimal threshold of zero in the threshold list. This guarantees that, in the worst case, all objects are included in the final attempt, ensuring the feasibility of reaching $\bm{SG}^{*}$ remains unchanged while significantly improving planning efficiency when accurate importance predictions are available.

\section{Experiment}\label{sec:experiments}
\subsection{Benchmarks}\label{subsection:benchmarks}
\subsubsection{B1-Tower Construction} In this benchmark, the robot is tasked with stacking a set of blocks into a specific configuration, starting from a random initial layout. The robot must compute a feasible task and motion plan using the available primitive actions: \textit{pick}, \textit{place}, \textit{stack}, and \textit{unstack}.

\subsubsection{B2-Tool Use} This task extends the \textit{B1} scenario by introducing a tool that enables the robot to \textit{pull} objects located outside its immediate workspace. The robot must autonomously decide when and how to use the tool, integrating this reasoning into its planning process. In addition to the stacking objective, the robot must satisfy inverse kinematics constraints and avoid collisions between the tool and surrounding objects. 

\subsubsection{B3-Cooking} In this benchmark, a Franka Emika robot arm mounted on a mobile base (with movement allowed along the Y-axis) is tasked with preparing a meal by sequentially washing, cutting, and cooking ingredients—each represented by colored blocks. The environment includes clutter in the form of transparent red blocks, simulating obstacles in a typical kitchen, which introduces additional collision constraints and limits accessibility to certain ingredients. The robot must also obey temporal constraints, such as processing multiple ingredients in a predefined order. 

Available actions include: \textit{pick}, \textit{place}, \textit{wash}, \textit{cut}, \textit{cook}, and \textit{cook-after}, where \textit{cook-after} enforces specific temporal dependencies across different ingredients.

\begin{table*}[thb!]
    \centering
    \begin{subtable}{\linewidth}
        \centering
        \label{subtable:block_complexVSdecomp}
        \setlength{\tabcolsep}{3.5pt} % Reduce horizontal padding
        \renewcommand{\arraystretch}{1.1} % Adjust row height for better readability
        \caption{Planning complexity vs. decomposability in \textit{Tower Construction}}
        \begin{tabular}{|c|c|c|c|c|c|c|c|c|c|c|c|}
        \hline
        \multicolumn{3}{|c|}{} & \multicolumn{3}{c|}{\textbf{4 Blocks}} & \multicolumn{3}{c|}{\textbf{6 Blocks}} & \multicolumn{3}{c|}{\textbf{8 Blocks}}\\
        \cline{4-12}
        \multicolumn{3}{|c|}{} & \textbf{Goal 0} & \textbf{Goal 1} & \textbf{Goal 2} & \textbf{Goal 0} & \textbf{Goal 1} & \textbf{Goal 2} & \textbf{Goal 0} & \textbf{Goal 1} & \textbf{Goal 2} \\
        \hline
        \multicolumn{3}{|c|}{\textbf{Time [s]}} & 0.2 $\pm$ 0.0 & 0.4 $\pm$ 0.0 & 2.9 $\pm$ 0.4 & 0.3 $\pm$ 0.02 & 3.2 $\pm$ 1.3 & 6.2 $\pm$ 0.8 & 0.5 $\pm$ 0.0 & 179.1 $\pm$ 0.8 & $\geq 180$ \\
        % \hline
        \multicolumn{3}{|c|}{\textbf{Horizon}} & 6 $\pm$ 0 & 7.6 $\pm$ 1.8 & 11.6 $\pm$ 1.7 & 10 $\pm$ 0 & 19.7 $\pm$ 2.8 & 18.6 $\pm$ 2.2 & 14 $\pm$ 0 & 24.5 $\pm$ 1.7 & 29.8 $\pm$ 3.2 \\
        % \hline
        \multicolumn{3}{|c|}{\textbf{Subgoals}} & 0 & 2 & 4 & 0 & 4 & 6 & 0 & 6 & 8 \\
        \hline
    \end{tabular}
    \end{subtable}
    
    \begin{subtable}{\linewidth}
        \centering
        \label{subtable:kit_complexVSdecomp}
        \setlength{\tabcolsep}{3.5pt} % Reduce horizontal padding
        \renewcommand{\arraystretch}{1.1} % Adjust row height for better readability
        \caption{Planning complexity vs. decomposability in \textit{Cooking}}
        \begin{tabular}{|c|c|c|c|c|c|c|c|c|c|c|c|}
        \hline
        \multicolumn{3}{|c|}{} & \multicolumn{3}{c|}{\textbf{2 Ingredients}} & \multicolumn{3}{c|}{\textbf{3 Ingredients}} & \multicolumn{3}{c|}{\textbf{4 Ingredients}}\\
        \cline{4-12}
        \multicolumn{3}{|c|}{} & \textbf{Goal 0} & \textbf{Goal 1} & \textbf{Goal 2} & \textbf{Goal 0} & \textbf{Goal 1} & \textbf{Goal 2} & \textbf{Goal 0} & \textbf{Goal 1} & \textbf{Goal 2} \\
        \hline
        \multicolumn{3}{|c|}{\textbf{Time [s]}} & 3.4 $\pm$ 0.3 & 3.1 $\pm$ 0.1 & 3.1 $\pm$ 0.1 & 14.5 $\pm$ 1.8 & 26.5 $\pm$ 14.5 & 25.7 $\pm$ 13.5 & $\geq 180$ & $\geq 180$ & $\geq 180$ \\
        % \hline
        \multicolumn{3}{|c|}{\textbf{Horizon}} & 10.6 $\pm$ 0.7 & 9.4 $\pm$ 1.3 & 9.4 $\pm$ 1.3 & 14.7 $\pm$ 1.0 & 15.2 $\pm$ 0.9 & 14.3 $\pm$ 1.6 & 20.0 $\pm$ 0.6 & 20.6 $\pm$ 1.4 & 21.8 $\pm$ 2.6  \\
        % \hline
        \multicolumn{3}{|c|}{\textbf{Subgoals}} & 3 & 3 & 3 & 3 & 4 & 5 & 4 & 6 & 7 \\
        \hline
    \end{tabular}
    \end{subtable}
    \caption{Planning complexity vs. decomposability in \textit{Tower Construction} and \textit{Cooking} benchmarks. The 'N/A' symbol indicates tasks where the solution time exceeded 180 seconds, reaching a time-out. Generally, task goals with a higher number of subgoals lead to increased planning horizon and time, but they also exhibit greater decomposability, showing the significance of goal decomposition.}
    \label{tab:complexVSdecomp}
    \vspace{-0.5cm}
\end{table*}

\subsection{Planning Complexity \& Decomposability of Task Goals}\label{subsec:complexVSdecomp}
In this section, we highlight the importance of goal decomposition by analyzing the relationship between a task goal’s decomposability and its planning complexity. We constructed a dataset of up to 40 demonstrations using PDDLStream \cite{garrett2020pddlstream} for two benchmark tasks: \textit{B1 – Tower Construction} and \textit{B3 – Cooking}. For each benchmark, we consider three distinct task goals, each exhibiting varying levels of complexity in terms of subgoal sequences. For example, in benchmark \textit{B1}, the task goals are defined as follows:
\begin{itemize}
    \item \textit{Goal 0} requires placing all blocks on the table, therefore has no intermediate subgoals.
    \item \textit{Goal 1} involves forming two distinct stacks of blocks in a specific order, such as \textit{(on A/C, B/D), (on-table B/D)}, therefore has two subgoal sequences.
    \item \textit{Goal 2} requires building a single stack in a specific order, such as \textit{(on A/B/C, B/C/D), (on-table D)}, thus must be achieved following one subgoal sequence.
\end{itemize}
We define planning complexity as the average time required to compute a feasible plan for a given task goal, and decomposability as the number of subgoals that can be identified to achieve that goal. We compare the average planning time, average planning horizon (i.e., the number of actions), and the number of subgoals by running PDDLStream across three different task goals and numbers of objects. A summary of these comparisons is presented in Table~\ref{tab:complexVSdecomp}.

Our results show that, with the same number of objects and similar planning horizons, planning time increases as the number of subgoals rises. This suggests that goals with more subgoals exhibit higher planning complexity—yet also greater decomposability. For example, in Table \ref{tab:complexVSdecomp}b, the planning time for reaching \textit{Goal 1} and \textit{Goal 2} increases substantially with the number of subgoals, when cooking a meal with three ingredients. Moreover, while cooking with four ingredients takes over 180 seconds for all task goals, the most complex task goal (\textit{Goal 2}) shows a lower success rate in our evaluations, reflecting the higher average planning time required.

More critically, tasks with the same number of objects but more complex goals generally demand a longer planning horizon and time. As shown in Table \ref{tab:complexVSdecomp}a, \textit{Goal 2} consistently requires more planning time and a longer horizon across different object counts compared to \textit{Goal 1}, and considerably more than \textit{Goal 0}. However, the number of subgoals also increases correspondingly with the increase of the planning horizon and time, indicating higher decomposability for these complex task goals. Interestingly, although placing eight blocks on the table (\textit{Goal 0} with \textit{8 Blocks}) is not decomposable and requires relative long planning steps (14) due to the large number of objects, it requires significantly less planning time compared to the other two more complex task goals when involving the same number of objects. This shows that tasks with subgoals are often harder to solve, thus demonstrating the necessity of our proposed goal decomposition.

In conclusion, the proposed goal decomposition method is particularly effective at breaking down complex task goals that require passing through multiple critical states in specific sequential order, which would otherwise lead to long planning time. The more subgoals a task goal has, the more complex it becomes, requiring longer planning time. However, this also increases its decomposability, as each state can serve as a checkpoint, dividing the task into smaller, more manageable subtasks.

\subsection{Replanning Efficiency of Our Approach}\label{subsec:reactive_tamp}
\begin{table*}[thb!]
    \centering
    \setlength{\tabcolsep}{3.5pt} % Reduce horizontal padding
    \renewcommand{\arraystretch}{1.1} % Adjust row height for better readability
    \begin{tabular}{|c|c|c|c|c|c|c|c|c|c|c|c|}
        \hline
        \multicolumn{3}{|c|}{} & \multicolumn{3}{c|}{\textbf{Tower Construction-Obj8}} & \multicolumn{3}{c|}{\textbf{Tool Use-Obj8}} & \multicolumn{3}{c|}{\textbf{Cooking-Obj4}}\\
        \cline{4-12}
        \multicolumn{3}{|c|}{} & \textbf{L1} & \textbf{L2} & \textbf{L3} & \textbf{L1} & \textbf{L2} & \textbf{L3} & \textbf{L1} & \textbf{L2} & \textbf{L3} \\
        \hline
        \multicolumn{3}{|c|}{\textbf{PDDLStream}} & N/A & N/A  & N/A  & N/A & N/A  & N/A  & $108.8 \pm 27.6$ & N/A  & N/A  \\
        % \hline
        \multicolumn{3}{|c|}{\textbf{LogicLfD}} & $6.4 \pm 0.6$ & $73.7 \pm 43.5$ & $115.8 \pm 29.1$ & $18.4 \pm 0.6$ & $62.5 \pm 24.2$ & $58.0 \pm 23.0$ & $165.8 \pm 5.5$ & N/A & N/A \\
        % \hline
        \multicolumn{3}{|c|}{\textbf{LogicLfD+}} & $4.5 \pm 0.4$ & $96.9 \pm 53.0$ & $127.3 \pm 79.6$ & $20.1 \pm 4.4$ & $76.2 \pm 50.2$ & $74.2 \pm 46.0$ & $38.7 \pm 6.0$ & $99.3 \pm 26.6$ & $99.5 \pm 26.7$ \\
        % \hline
        \multicolumn{3}{|c|}{\textbf{Ours No OR}} & $4.3 \pm 0.5$ & $46.6 \pm 38.8$ & $62.7 \pm 55.5$ & $22.8 \pm 6.3$ & $79.3 \pm 23.3 $ & $58.0 \pm 23.0$ & $60.0 \pm 24.4$ & $118.1 \pm 10.4$ & $116.0 \pm 4.7$ \\
        % \hline
        \multicolumn{3}{|c|}{\textbf{Ours}} & $\bm{2.4 \pm 0.6}$& $\bm{7.5 \pm 0.9}$ & $\bm{7.1 \pm 0.8}$ & $\bm{7.2 \pm 0.9}$ & $\bm{11.7 \pm 1.0} $ & $\bm{12.7 \pm 0.6}$ & $\bm{24.3 \pm 4.6}$ & $\bm{34.1 \pm 7.5}$ & $\bm{34.5 \pm 7.8}$ \\
        \hline
    \end{tabular}
    \caption{Comparison of total replanning time $[s]$ from a randomly disturbed state to the task goal across three benchmarks. 'N/A' indicates no successful trials within the 180-second timeout. Results demonstrate that \textit{Ours} achieves significantly better reactivity and efficiency under various disturbances compared to the baseline methods because of the integration of problem decomposition.}
    \label{tab:reactive}
    % \vspace{-0.5cm}
\end{table*}

This section evaluates the reactivity of our proposed replanning method by comparing it against several baselines: \textit{PDDLStream}, \textit{LogicLfD} \cite{zhang2024logic}, \textit{LogicLfD$^{+}$}, and our method without object reduction (denoted as Ours No OR). \textit{LogicLfD} takes a single demonstration and finds the temporally closest seen symbolic state and the plan to reach this state simultaneously by running the \textit{PDDLStream} solver to reach all symbolic states on the demonstrated trajectory concurrently, then considers the one connected with minimal time as the closest state. \textit{LogicLfD$^{+}$} is a variant that replaces the original demonstration with our discovered subgoal sequences, thereby incorporating goal decomposition and a model-based \textcolor{blue}{computational} distance, but without any object reduction. To isolate the effect of each module, we also evaluate our full method with goal decomposition and \textcolor{blue}{computational} distance estimation but without object reduction. This variant helps highlight the efficiency gains contributed by the object reduction mechanism. The following types of online disturbances are considered:
\begin{itemize}
    \item \textit{L1 Disturbance:} Changes in object configurations that require replanning and subgoal update.
    \item \textit{L2 Disturbance:} Addition of environmental objects that can be ignored during replanning. 
    \item \textit{L3 Disturbance:} Introduction of additional objects that requires consideration during replanning.
\end{itemize}

For evaluation, we randomly initialize the environment and introduce disturbances before completing the target task. It is important to note that recovering from disturbances \textcolor{blue}{may require not only replanning to the closest subgoal but also to subsequent subgoals}. To ensure a fair comparison, we measure the total replanning time from the disturbed state to the final task goal. The tasks conducted in the \textit{Tower Construction} and \textit{Tool Use} domains involves eight blocks (denoted as \textit{Obj8} in the table) under \textit{L1} disturbances and three additional blocks under \textit{L2} and \textit{L3} disturbances, targeting task \textit{Goal 2}. For the \textit{Cooking} tasks, we used four ingredients (denoted as \textit{Obj4}) under \textit{L1} disturbances, with three additional blockers introduced for \textit{L2} and \textit{L3} disturbances. The results are summarized in Table~\ref{tab:reactive}. 

Overall, we observed that vanilla \textit{PDDLStream} struggles to efficiently handle disturbances, achieving a success rate of only $60\%$ under \textit{L1} disturbances in the \textit{Cooking} domain and failing all tasks in \textit{Tower Construction} and \textit{Tool Use} within the 180-second timeout. \textit{LogicLfD} improves upon \textit{PDDLStream} by considering symbolic states on the demonstrated trajectory as subgoals. However, its replanning efficiency is heavily dependent on the quality of the demonstrated trajectory, leading to instability. For instance, in the Cooking domain, a single demonstration proves insufficient for successful replanning within the timeout. In contrast, our generated subgoal sequences guide the TAMP solver toward only the \textit{necessary} subgoals, making \textit{LogicLfD$^{+}$} more consistent and reliable than LogicLfD. Since our learned \textcolor{blue}{computational} distance is an approximation, it may occasionally select the second closest subgoal, resulting in slightly longer replanning time than a model-based \textcolor{blue}{computational} distance definition in \textit{LogicLfD$^{+}$}. However, this model-based approach requires reasoning over all objects, limiting efficiency. In our full method \textit{Ours}, we leverage the approximate \textcolor{blue}{computational} distance to identify the closest subgoal, then apply object reduction to significantly improve replanning efficiency. As demonstrated in Table \ref{tab:reactive}, our approach consistently outperforms all baselines, achieving faster and more reliable replanning across domains.

In addition, we directly applied the subgoal sequences and GNN model trained on the \textit{Tower Construction} domain to the \textit{Tool Use} domain to evaluate their robustness in the presence of new actions and objects. \textcolor{blue}{This is achieved by simply augmenting the original object set with \texttt{(tool, hook)}}. We observed that the subgoal sequences consistently guided the TAMP solver during replanning, while the GNN model reliably identified the temporally closest subgoals in the new domain. Notably, the object reduction strategy remained effective: it consistently minimized the number of objects considered for subgoals that did not require the use of the \textit{hook}, while automatically reverting to full-environment planning for subgoals involving tool usage. As a result, we observed an increase in planning time in the \textit{Tool Use} domain compared to \textit{Tower Construction}, primarily due to the inclusion of all environmental objects when tool-based actions were necessary. These results demonstrate that the learned subgoal sequences and trained GNN model are transferable to structurally similar domains that introduce new objects or extend the action set. This generalization capability is crucial, as it allows for the continuous accumulation of domain-specific experience and incremental refinement of the subgoal sequences and GNN model in a lifelong learning setting—without altering the overall architecture of the proposed replanning framework.

Additional robustness analyses and ablations of our problem decomposition modules are provided in Appendix~\ref{app:robustness}, where we examine why and when these modules can be reused to solve new tasks.

\section{From blocks to Realistic Kitchen Scenarios}\label{sec:exp_realistic}
\subsection{PR2 Whole-body Planning in PyBullet}
\begin{figure*}[!thb]
    \centering
    \begin{subfigure}{0.48\linewidth}
        \centering
        \includegraphics[width=\linewidth]{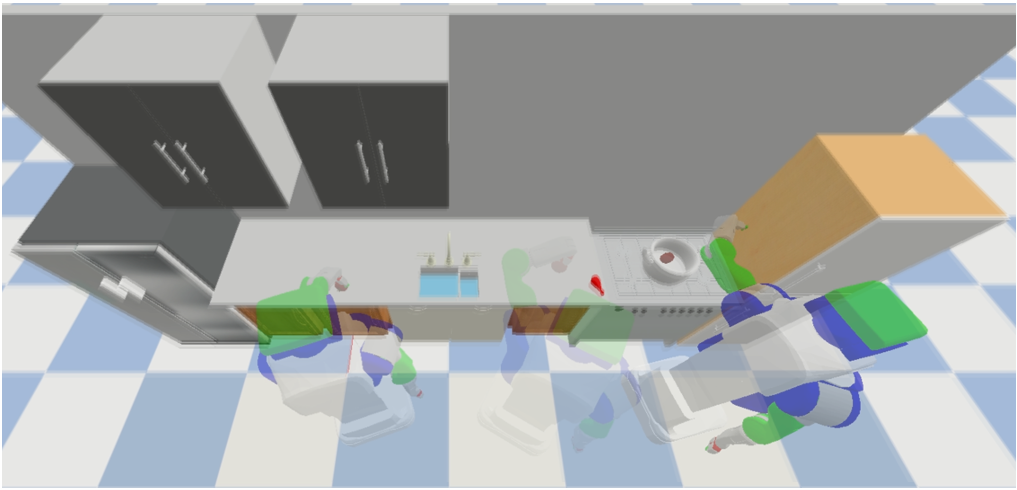}
        \caption{}
        \label{fig:pr2_kitchen_b}
    \end{subfigure}
    \hspace{0.02cm}
    \begin{subfigure}{0.48\linewidth}
        \centering
        \includegraphics[width=\linewidth]{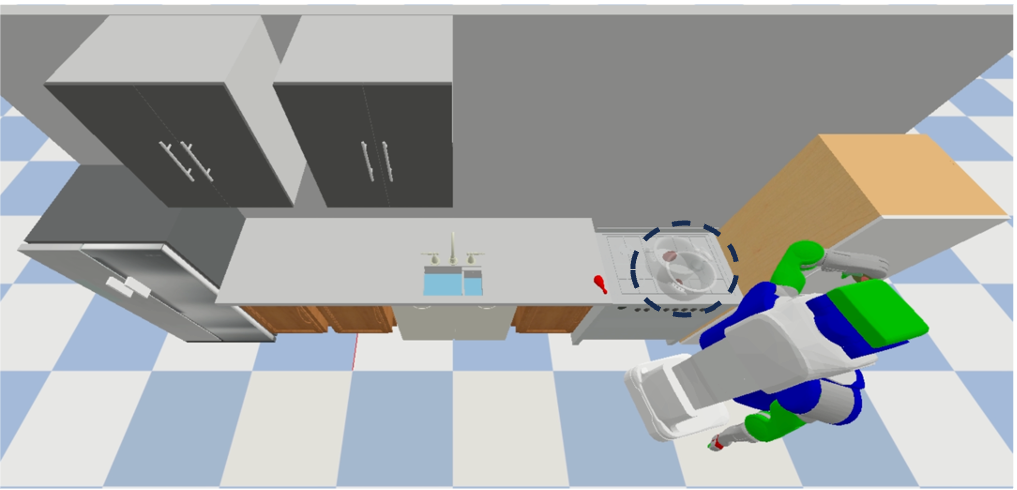}
        \caption{}
        \label{fig:pr2_kitchen_c}
    \end{subfigure}
    \\
    \begin{subfigure}{0.48\linewidth}
        \centering
        \includegraphics[width=\linewidth]{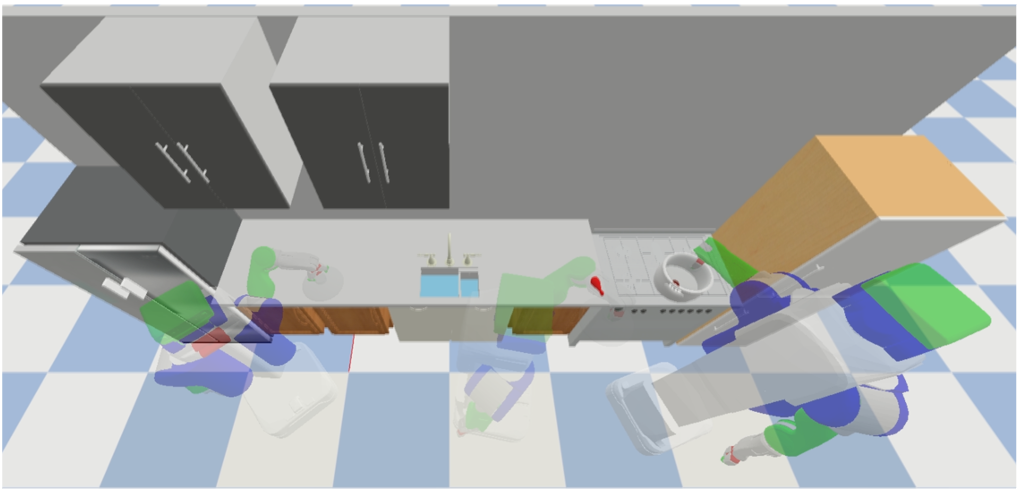}
        \caption{}
        \label{fig:pr2_kitchen_d}
    \end{subfigure}
    \hspace{0.02cm}
    \begin{subfigure}{0.48\linewidth}
        \centering
        \includegraphics[width=\linewidth]{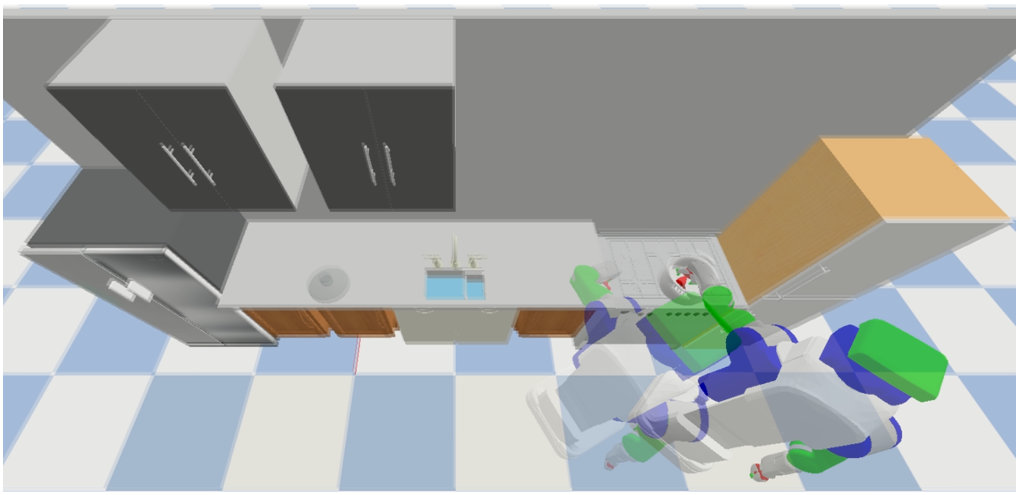}
        \caption{}
        \label{fig:pr2_kitchen_e}
    \end{subfigure}
    \caption{Application of our efficient replanning approach for a PR2 robot cooking a meal with a tomato and a chicken leg, under L3 disturbances. (a) The robot picks and cooks the tomato; (b) L3 disturbances occur (black dash line), including a lid being placed on the pot and interference with the tomato; (c) The robot reacts to the disturbance by removing the lid from the pot; (d) The robot picks and cooks the chicken leg, successfully completing the preparation of the target meal.}
    \label{fig:reactive_tamp_pr2}
    % \vspace{-0.5cm}
\end{figure*}

\begin{figure*}[thb]
    \centering
    \begin{subfigure}{0.16\linewidth}
        \centering
        \includegraphics[width=\linewidth]{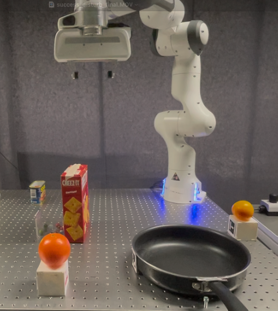}
        \caption{}
        \label{fig:real_kitchen_a}
    \end{subfigure}
    \begin{subfigure}{0.16\linewidth}
        \centering
        \includegraphics[width=\linewidth]{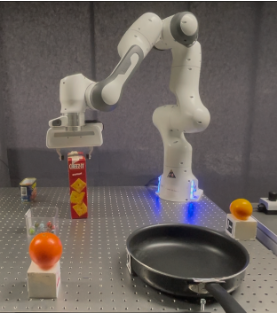}
        \caption{}
        \label{fig:real_kitchen_b}
    \end{subfigure}
    \begin{subfigure}{0.16\linewidth}
        \centering
        \includegraphics[width=\linewidth]{images/pick_place_rice_box.png}
        \caption{}
        \label{fig:real_kitchen_c}
    \end{subfigure}
    \begin{subfigure}{0.16\linewidth}
        \centering
        \includegraphics[width=\linewidth]{images/add_disturbance.png}
        \caption{}
        \label{fig:real_kitchen_d}
    \end{subfigure}
    \begin{subfigure}{0.16\linewidth}
        \centering
        \includegraphics[width=\linewidth]{images/remove_disturbance.png}
        \caption{}
        \label{fig:real_kitchen_e}
    \end{subfigure}
    \begin{subfigure}{0.16\linewidth}
        \centering
        \includegraphics[width=\linewidth]{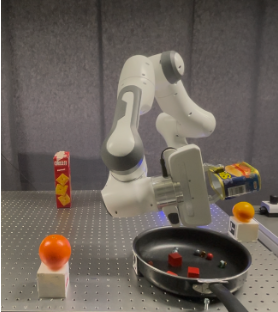}
        \caption{}
        \label{fig:real_kitchen_f}
    \end{subfigure}
    \caption{Full screenshots of cooking a meal using a Franka arm under L1 and L2 disturbances in real world. (a) Initial environment state; (b) The robot removes the crackers box to enable feasible tomato box grasping; (c) The robot picks and pours the tomato box into the pot for cooking; (d) Disturbances (black dash line) are introduced by repositioning the meat can and the crackers box; (e) The robot reacts to the disturbance by removing the crackers box; (f) The robot picks and pours the meat from the can into the pot, completing the preparation of the target meal.}
    \label{fig:reactive_tamp_franka}
    % \vspace{-0.5cm}
\end{figure*}

In earlier benchmarks, we modeled ingredients as blocks to simplify perception and manipulation. Here, we lift that simplification and evaluate our replanning approach in a realistic kitchen scenario in PyBullet \cite{coumans2021pybullet, yang2023sequence}. The PR2 is tasked with preparing a soup by processing a tomato and a chicken leg while avoiding collisions between all environment objects and its whole body. Importantly, the task specification provides only the final goal and safety constraints—it does not encode any temporal ordering between ingredients. The desired precedence—cook the tomato before the chicken—is present only implicitly in the demonstrations derived from the recipe. The robot must therefore infer the cooking sequence from the demonstrations and execute a plan that respects this ordering.

Specifically, we expect a PR2 robot with an extended base capable of changing its height is to cook the tomato and chicken to make a soup. For planning in this domain, we augmented the state abstraction in \textit{B3-Cooking} by including kitchen objects under unseen object type such as doors and design a new set of action abstraction where PR2 plans both base and arm motions for cooking meals. It does not have the knowledge that the pot needs to be \textit{clear} before cooking any ingredients with it. Therefore, the robot needs to try to \textit{pick} and \textit{place} different movable objects for finding a collision-free task and motion plan to achieve the target goal. This experiment thus serves as an evaluation of the reactivity and robustness of the proposed method under L2 and L3 disturbances.

Thanks to the robustness of our problem decomposition, we managed to directly apply the subgoal sequences and trained GNN from \textit{B3-Cooking} in this new task (no retraining). The newly added cabinets and fridge constitute an L2 disturbance (additional environmental objects). We also introduce an L3 disturbance by placing a lid on the pot after the robot finishes cooking the tomato. This change forces the PR2 to replan the remaining steps for cooking the chicken. Using the same subgoal sequences and GNN, the solver finds a feasible plan by (i) correctly predicting the temporally nearest subgoal and (ii) marking all environmental objects as essential during object reduction. The mean replanning time is 76 s, and replanning process is shown in Figure~\ref{fig:reactive_tamp_pr2}. 

In this task, we found that two modules dominate the replanning time: (i) generating collision-free, stable grasp poses and (ii) whole-body motion planning. Consequently, replanning latency remains high even with our problem decomposition. As future work, we plan to incorporate learning-based grasp synthesis and learning-guided motion planning to warm-start or accelerate these subproblems, with the goal of achieving near real-time task-and-motion replanning.

\subsection{Real-world Cooking with Franka Robot Arm}

We then conducted a real-world meal cooking experiment using a 7-DoF Franka robot arm equipped with a RealSense D415 camera fixed in the environment to capture the poses of environmental objects marked with visual markers, as shown in Figure~\ref{fig:reactive_tamp_franka}. For this experiment, we utilized the subgoal sequences and the GNN model derived from \textit{B3-Cooking} with objects in the YCB dataset. The task is simplified by assuming ingredients are pre-washed and pre-cut, ready to be cooked directly if a feasible grasping and pouring for cooking path is found. However, challenges in non collision-free grasping due to blockers remain. We introduced an orange and a peach as additional ingredients that are non-essential for cooking the target meal in the initial environment configuration, corresponding to L2 disturbances. Additionally, we replaced the crackers box to obstruct the robot from cooking the meat in the can, as an L1 disturbance. The newly added orange and peach are never seen object type in the original tasks and used GNN model. We observed that our method found feasible plans for reacting to disturbances by predicting the correct subgoal and overall considering all the environmental objects as essential when disturbing the crackers box to block the manipulation of ingredients (taking around 2.87 seconds). However, they still managed to filter out non-essential objects when the crackers box does not block the grasping of essential ingredients and react to disturbances within around 0.83 second. Figure~\ref{fig:reactive_tamp_franka} shows representative screenshots of the task execution. 

Please refer to our supplementary video in our project website for more details and videos about our simulated and real-world experiments.

\section{Discussions and Limitations}
\textbf{Simplified Motion Planning}: In this paper, we primarily focus on simple motion planning problems, such as picking and placing objects, in order to isolate and study the influence of planning horizon length and the number of environment objects on planning efficiency. For the tasks in Tables~\ref{tab:complexVSdecomp} and~\ref{tab:reactive}, we define motion planning streams that can be solved in approximately 3 milliseconds. However, when the constraint graph search involves more than eight objects and a planning horizon exceeding 30, the computation time can surpass 30 seconds. Our results show that the proposed approach reduces such planning time by decreasing both the planning horizon and the number of objects involved. Even for tasks with complex motion planning subproblems, we expect our method to remain effective by reducing the number of motion planning subproblems encountered during replanning through planning horizon reduction.

\textbf{Robustness of Our Problem Decomposition Modules}: In Appendix~\ref{app:robustness}, we numerically demonstrate the robustness of our goal decomposition, computational distance, and object reduction to domain variants, such as changes in object geometry, object numbers, robot numbers, action-abstraction hierarchy, and expanded action/skill sets. We show that our proposed problem decompositions are robust to domain variants as long as they do not change the implicit domain rules in the data-collection domain. In this paper, we leverage this robustness to generalize heuristics learned from simplified domains to accelerate online replanning in realistic domains (Section~\ref{sec:exp_realistic}). This robustness can also be helpful for distributed demonstration generation, where demonstrations for the same task goal are collected in different environment setups that share the same domain rules. 

\textbf{Consideration of Geometrical States in Goal Decomposition}: The learned subgoal sequences consider geometrical information from the demonstrations. In our goal decomposition, geometrical information (e.g., object poses) is encoded as logical states via geometrical predicates (e.g., \texttt{(atpose ?obj ?p)}). For two literals derived from geometrical predicates, we treat them as equal if they apply to the same object and their pose distance is within 1 cm (threshold). However, for the tasks we investigate, no specific poses need to be traversed in a particular order. Consequently, the generated subgoal sequences tend to include only logical information. As future work, we will investigate which tasks require geometrical information to be captured as subgoals, to better illustrate this property.

\textbf{Probabilistic Completeness}: Our object-reduction procedure preserves the probabilistic completeness (PC) of the base \textit{PDDLStream} solver.  In the worst case, it reverts to planning with the full set of environment objects; hence no feasible plan is ruled out, and the PC guarantee of the original solver is retained. However, without additional assumptions, the goal-decomposition step does \emph{not} preserve the PC guarantee of the original \textit{PDDLStream} solver in replanning mode. In principle, while pursuing a selected subgoal, a TAMP solver eventually enters irreversible intermediate states in which the selected subgoal is a subset; those additional states may render subsequent subgoals or the event task goal infeasible. For example, in the \textit{Cooking} domain with the recipe “cook tomato then meat,” starting from a random initial state where the closest subgoal is \texttt{(cooked tomato)}, a solver might reach \texttt{(cooked tomato)} and \texttt{(cooked meat)}, but with meat cooked \emph{before} tomato, violating the required order and causing task failure. In practice, we did not observe this failure mode: achieving unrelated literals typically incurs extra complexity and time, so solvers tend to avoid it naturally. A principled way to eliminate this risk is to run the TAMP solver in a minimal-time (or minimal-cost) mode and to prove that optimal policies avoid such detours; however, solving this optimal TAMP problem and establishing the corresponding PC result are left for future work.

\textbf{Approximated Computational Distance}: In addition, our current computational distance metric approximates the computational complexity of planning problems under the assumption that motion planning subproblems require similar planning time, which may not hold in complex scenes. Furthermore, during replanning, we currently consider only the computational time when selecting the closest subgoal, without accounting for actual execution time. In future work, it would be valuable to design improved computational distance metrics to enable more optimal subgoal selection. 

\textbf{Efficient Replanning with Feedforward and Feedback Terms}: Our current replanning strategy neither \emph{reuses} previously evaluated feasible motion-planning results nor explicitly reasons about failure cases; instead, it replans from scratch. Previous work has shown that LLMs can translate cooking recipes into symbolic task languages for planning \cite{sakib2024cooking, mavrogiannis2024cook2ltl} and analyze failure information to enable more informative replanning \cite{pmlr-v270-curtis25a}. Our goal-decomposition method generates subgoal sequences from few demonstrations, thus is complementary to LLMs and would be beneficial for online fine-tuning of the transferred results with online collected data. As future work, we will explore using LLMs as \emph{feedforward} and \emph{feedback} terms in our replanning approach to reuse existing instructions and past planning experience \emph{pre-failure}, moving toward real-time task-and-motion replanning.

\section{Conclusion}
In this work, we developed an efficient task and motion replanning approach that integrates our proposed problem decomposition learning method with classical TAMP solvers to enable efficient sequential multi-object manipulation in dynamic environments. We introduced techniques to learn subgoal sequences and \textcolor{blue}{computational} distance function from demonstrations, which significantly reduce the planning horizon and the number of objects involved during replanning, resulting in substantial improvements in replanning efficiency. Through extensive numerical experiments, we also showed that task goals with more subgoals generally require longer planning times but are inherently more decomposable. This key observation motivated the incorporation of goal decomposition and \textcolor{blue}{computational} distance learning to reduce the effective planning horizon, thereby enhancing the reactivity of TAMP solvers.

\bibliography{IEEEabrv,main}
\bibliographystyle{IEEEtran}
\clearpage

\appendix
\subsection{Pseudocode of Our Replanning Approach}\label{app:pseudocode}
In Algorithm~\ref{alg:reactive_tamp}, we provide the pseudocode of our efficient task-and-motion replanning approach. We assume the environment $env$ is known. This includes both the real-world environment and their models used for generating the next system state when the robot executes a planned action $\bm{a}$ and motion $\bm{\tau}$. We also assume that the task goal $\bm{G}$ and the corresponding subgoal sequences $\bm{SG}$, represented as a directed acyclic graph (DAG), are known. The GNN model is pre-trained offline to compute the computational distance $d(\bm{s}, \bm{SG}_i)$, and an importance threshold $\epsilon$ is specified for calculating the number of important objects. The outputs of our approach are a feasible action $\bm{a}$ and a motion plan $\bm{\tau}$, which guide the system towards the task goal $\bm{G}$ as efficiently as possible, even when disturbances occur.

\begin{algorithm}[t!hb]
    \caption{Efficient Task and Motion Replanning}
    \label{alg:reactive_tamp}
    \textbf{Given:} Environment $env$, Task Goal $\bm{G}$, Subgoal Sequence(s) $\bm{SG}$, Trained \textit{GNN} model, Importance Threshold $\epsilon$ \\
    \textbf{Output:} next action $\bm{a}$ and motion $\bm{\tau}$ \\
    \textbf{Initialization:} \\
        $\bm{s} = \textit{CurrentEnvironmentState}()$ \\
        % $\bm{SG}^{'} = Nodes(\bm{SG})$ \\
        % \textcolor{blue}{\# all subgoals is possible} 
    \While{$\bm{G} \not \in \bm{s}$}{
        % \textcolor{blue}{\# find the closest subgoal not reached yet} \\
        $\bm{SG}^{*} = \textit{ClosestSubgoalGen}(\bm{s}, \bm{SG}, \textit{GNN}, \epsilon)$ \\
        % \While{$d == 0$}{
        %     $\Tilde{\bm{G}'} = SubgoalsGen(\bm{G}', id)$ \\
        %     $id, t_d, \bm{v} = \textit{ClosestSubgoalGen}(\mathcal{S}_{C}, \Tilde{\bm{G}'}, \bm{T}_d, \epsilon)$ \\   
        % }
        % \textcolor{blue}{\# generate subproblems with different $\epsilon$} \\
        $\bm{P} = SubproblemGen(\bm{s}, \bm{SG}^{*}, \textit{GNN}, \epsilon)$ \\
        % \textcolor{blue}{\# concurrently solve subproblems} \\
        $P_L = SubproblemParallel(\bm{P})$ \\
        % \textcolor{blue}{\# generate action and plan} \\
        $\bm{a}, \bm{\tau} = \textit{ActionMotionGen}(\textit{env}, P_L, \bm{SG}^{*})$ \\
        $\Tilde{\bm{s}} = env(\bm{a}, \bm{\tau})$ \\
        $\bm{s} = \textit{CurrentEnvironmentState}()$ \\
    }
\end{algorithm}

Initially, all the nodes in the subgoal sequence(s) $\bm{SG}$ are considered as potential subgoals for identifying the closest one. The system continuously captures the current environment state $\bm{s}$ after executing each planned action and motion $\bm{a}, \bm{\tau}$ (lines 4 and 11). If the task goal $\bm{G}$ is not a subset of the current environment state $\bm{s}$, meaning the task goal has not yet been reached, the system proceeds to find the closest not-yet-reached subgoal $\bm{SG}^{*}$ using the \textit{ClosestSubgoalGen} function (lines 6). Next, a set of subproblems is generated with different object sets (line 7), and these subproblems are solved in parallel to find a feasible task and motion plan $P_L$ that reaches the closest subgoal $\bm{SG}^{*}$ (line 8). This function terminates all parallel processes as soon as $P_L$ is found, ensuring the plan is generated with as few objects—and thus as little planning time—as possible.

The robot then executes the first action and motion $\bm{a}, \bm{\tau}$ from $P_L$, and the system checks whether the resulting environment state $\Tilde{\bm{s}}$ matches the expected state $\bm{s}$. If there is a mismatch, indicating that the environment has been disturbed, the system re-runs the previous steps to find a new plan $P_L$. If the environment remains unchanged, the system continues executing the next action and motion from $P_L$ and updates $P_L$ iteratively until the current subgoal is reached or a disturbance occurs.

\subsection{Why and When Task Goals are Decomposable}\label{app:goal_decomposition}
In this section, we explain the relationship between goal decomposition and domain rules, and illustrate how this motivates the identification of necessary subgoals and the proposed goal decomposition method.

We assume that $\mathcal{D}$ is characterized by a known PDDL-based state abstraction, along with a set of conventions that convert infinite geometric states $g \in \mathcal{G}$ into finite symbolic states $s \in \mathcal{S}$. Domain rules $\mathcal{R}$ indicate the sequential orders of achieving some predicates, such as one predicate can only be achieved if another set of predicates are true beforehand, which can be applied on different objects as long as they are within the same object type. Here, we do not assume knowledge of these domain rules, but argue that their existence results in some essential \textit{literals} that must be achieved in sequence before reaching one specific \textit{literal} in the task goal $\bm{G}$. More importantly, the achievement of one \textit{literal} within task goal $\bm{G}$ usually influences the achievement of the rest, due to the presence of domain rules. Therefore, we aim to identify the coupled sequences of \textit{literals} for achieving the task goal $\bm{G}$, where all \textit{literals} should be reached simultaneously. We also argue that if we are capable of finding such \textit{literal} sequences for a task goal, we consider this task goal as decomposable. 

In the following, we provide more details about previous arguments by answering three essential questions:
\begin{itemize}
    \item[Q1] What are domain rules?
    \item[Q2] Why domain rules result in necessary \textit{literal} sequences for achieving one \textit{literal} in the task goal?
    \item[Q3] Why the above \textit{literal} sequences merge when multiple \textit{literals} should be achieved concurrently in the task goal?
\end{itemize}

\textit{Q1:} Domain rules may originate from various sources, such as physical laws, user specifications, etc. In planning domain $\mathcal{D}$, such domain rules can be described with predicates as:
\begin{equation}
    r_i := \psi_i \: \text{only if} \: \{\psi_{i, 1}, \psi_{i, 2}, \dots, \psi_{i, m} \}.
\end{equation}
where the $i$-th domain rule $r_i$ indicates that predicate $\psi_i$ is true only if $\{\psi_{i, 1}, \psi_{i, 2}, \dots, \psi_{i, m} \}$ are achieved beforehand. Such domain rules can be observed in lots of multi-object manipulation tasks. For example, in block stacking domain, one task rule $r_1:=$ $\psi_1$ \textit{if} $\{\psi_{1,1}$, $\psi_{1,2}\}$ states that cube A can be \textit{on} cube B only if cube B is \textit{clear} and stably located on a surface in the environment beforehand, due to physical stability consideration. Here $\psi_1=$ \textit{(on, A, B)}, $\psi_{1,1}=$ \textit{(clear, B)}, and $\psi_{1,2}=$ \textit{(on, B, X)} where X can be any surface in the environment. In kitchen cooking applications, another possible domain rule $r_1':=$ $\psi_1'$ \textit{if} $\{\psi_{1,1}'$, $\psi_{1,2}'\}$ states that ingredient A can be \textit{cooked} only if it has been \textit{washed} and \textit{cut} beforehand, due to user specifications, even if it is physically feasible to make A cooked without washing or cutting it first. In this case, $\psi_1'=$ \textit{(cooked, A)}, $\psi_{1,1}'=$ \textit{(washed, A)}, and $\psi_{1,2}'=$ \textit{(cut, A)}.

\textit{Q2:} Moreover, such domain rules may depend on each other, resulting in sequences of predicates that must be achieved before reaching a target goal predicate. For example, in the kitchen domain, we have the domain rule $r_1':=$ $\psi_1'$ \textit{only if} $\{\psi_{1,1}'$, $\psi_{1,2}'\}$ for achieving $\psi_1'=$ \textit{(cooked, A)} and another domain rule $r_2':=$ $\psi_{1,1}'$ \textit{only if} $\psi_{1,2}'$ which indicates that an ingredient can only be cut if it has been washed beforehand. Therefore, to reach the task goal $\psi_1'=$ \textit{(cooked, A)}, one must compose the corresponding domain rules, which can be formally summarized as:
\begin{equation}
    \begin{split}
        r_1'&:=\psi_1'\: \textit{only if} \: \{\psi_{1,1}', \psi_{1,2}'\} \\
        r_2'&:=\psi_{1,1}' \: \textit{only if} \: \{\psi_{1,2}'\} \\
    \end{split}
\end{equation}
\begin{equation}
    % \bm{G} \Longrightarrow r_1' \oplus r_2':= \psi_1'\: \textit{if} \: \{\psi_{1,1}'\} \: \textit{if} \: \{\psi_{1,2}'\} \\
    r_1' \oplus r_2':= \psi_1'\: \textit{only if} \: \{\psi_{1,1}'\} \: \textit{only if} \: \{\psi_{1,2}'\} \\
\end{equation}
Correspondingly, we will always observe the robot first reaching $\psi_{1,2}'=$\textit{(washed, A)}, then reaching $\psi_{1,1}'=$\textit{(cut, A)}, and eventually $\psi_1'=$ \textit{(cooked, A)}, regardless of the initial states. 

\textit{Q3:} Besides, a complex long-horizon planning problem typically involves a task goal that is the conjunction of several \textit{literals}. For example, a task goal $\bm{G}$ in a block stacking domain can be to achieve both $\psi_1=$ \textit{(on, A, B)} and $\psi_2=$ \textit{(on, B, C)}. Given the same domain rule $r_1$, we will need to reach $\psi_{1,1}=$ \textit{(clear, B)} and $\psi_{1,2}=$ \textit{(on, B, X)} before achieving $\psi_1$, as well as $\psi_{2,1}=$ \textit{(clear, C)} and $\psi_{2,2}=$ \textit{(on, C, X)} before achieving $\psi_2$. Since $\psi_{1,2}=$ \textit{(on, B, X)} is expected to be $\psi_2=$ \textit{(on, B, C)} in the task goal specification, the essential \textit{literal} sequences for $\psi_1$, $\psi_2$ need to be merged because the coupling effect of goal \textit{literals}:
\begin{equation}
    \begin{split}
        r_1^{1}&:=\psi_1 \: \textit{only if} \: \{\psi_{1,1}, \psi_{1,2}(X)\} \\
        r_1^{2}&:=\psi_2 \: \textit{only if} \: \{\psi_{2,1}, \psi_{2,2}(X)\} \\
    \end{split}
\end{equation}
\begin{equation}
    % \bm{G} \Longrightarrow r_1^{1} \oplus r_1^{2}:= \{\psi_1\} \: \textit{if} \:  \{\psi_{1,1}, \psi_2\} \: \textit{if} \:  \{\psi_{2,1}, \psi_{2,2}\} 
    r_1^{1} \oplus r_1^{2}:= \{\psi_1\} \: \textit{only if} \:  \{\psi_{1,1}, \psi_2\} \: \textit{only if} \:  \{\psi_{2,1}, \psi_{2,2}\} 
\end{equation}
where $r_1^{1}, r_1^{2}$ indicate two different instances of the same domain rule $r_1$. With such rule composition, to achieve task goal $\bm{G}=\{\psi_1, \psi_2\}$, we will always observe the robot first reaching \textit{(on, C, X)} \textit{and} \textit{(clear, C)}, then \textit{(on, B, C)} \textit{and} \textit{(clear, B)}, and eventually \textit{(on, A, B)}, regardless of the initial states.

Therefore, we argue that if there are a set of such domain rules, we will observe some essential \textit{literals} that are achieved in sequence before achieving the task goal. Since no matter which action plan we eventually find as feasible with TAMP solvers, the system has to pass through such essential \textit{literals} in order, we can thus use such \textit{literal} sequences as subgoals for TAMP solvers to reach goal $\bm{G}$. Task goal $\bm{G}$ is thus considered as decomposable. 

Note that subgoals are composed of \textit{literals} that consistently appear in a specific order across past experiences, representing \textit{partial} states of environmental objects at each scene. Each essential subgoal therefore corresponds to a \textit{partial} state of the entire set of environmental objects. Importantly, to determine if a task goal is decomposable, we do not require any knowledge of such domain rules and its existences. Instead, we can run our proposed goal decomposition method to identify if such sequences of subgoals exist from past experiences and based on such existence to evaluate the decomposability of the task goal and the existence of such domain rules.

In classical TAMP problems, we typically assume access to an action abstraction that explicitly encodes domain rules via preconditions. In this paper, we do not assume such an abstraction during the learning phase; instead, we aim to discover these domain rules from state-only demonstrations. Correspondingly, our learned subgoal sequences show robustness to the choice of action abstraction (shown in follow-up section). Moreover, some goal-specific domain rules may not be explicitly included in the action abstraction. For example, in the \textit{Cooking} domain, the abstracted \textit{cooking} action should not encode the specific order of processing different ingredients, as this is a goal-specific preference. With our goal-decomposition approach, we discover such temporal rule for the specific task goal from demonstrations. These properties enable our goal decomposition to generate action-abstraction-agnostic subgoal sequences without explicit knowledge of domain rules, thereby enhancing its generalization and robustness.

\subsection{Robustness of Our Problem Decomposition Modules}\label{app:robustness}
In Section~\ref{subsec:reactive_tamp}, we observed that the subgoal sequences and trained GNN model from \textit{B1-Tower Construction} can be directly transferred to \textit{B2-Tool Use} domain with unseen object types and actions (new tool object type and tool use actions). In this section, we take a deeper look at the robustness of each problem decomposition module, aiming at explaining why and when they show such capability.

\begin{table*}[thb!]
    \centering
    \tiny
    \setlength{\tabcolsep}{2.5pt} % Reduce horizontal padding
    \renewcommand{\arraystretch}{1.1} % Adjust row height for better readability
    \begin{subtable}{\linewidth}
        \caption{Robustness of Subgoal Sequence(s) from \textit{B1-Tower Construction}}
        \label{subtable:block_goal_generalize}
        \centering
        \begin{tabular}{|c|c|c|c|c|c|c|c|c|c|c|c|c|c|c|c|c|}
            \hline
            \multicolumn{2}{|c|}{} & \multicolumn{3}{c|}{\textbf{Object Geometry-Block6}} & \multicolumn{3}{c|}{\textbf{Number of Objects-Block6}} & \multicolumn{3}{c|}{\textbf{Number of Robots-Block4}} & \multicolumn{3}{c|}{\textbf{Action Abstration-Block4}} & \multicolumn{3}{c|}{\textbf{Action Set-Block8}}\\
            \cline{3-17}
            \multicolumn{2}{|c|}{} & \textbf{Goal0} & \textbf{Goal1} & \textbf{Goal2} & \textbf{Goal0} & \textbf{Goal1} & \textbf{Goal2} & \textbf{Goal0} & \textbf{Goal1} & \textbf{Goal2} & \textbf{Goal0} & \textbf{Goal1} & \textbf{Goal2} & \textbf{Goal0} & \textbf{Goal1} & \textbf{Goal2} \\
            \hline
            \textbf{Success Rate} & \textit{PDDLStream} & 100 & 100 & 100  & 100 & 100 & 100  & 100 & 100 & 100  & 100 & 100 & 100  & 100 & 90 & 0 \\
            % \cline{2-17}
            & \textit{PDDLStream}${+}$ & 100 & 100 & 100  & 100 & 100 & 100 & 100 & 100 & 100 & 100 & 100 & 100 & 100 & 100 & 100\\
            \cline{1-2}
            \textbf{Time} [\textit{s}] & \textit{PDDLStream} & 0.4 $\pm$ 0.0 & 1.0 $\pm$ 0.2 & 8.3 $\pm$ 6.0 
            & 7.8 $\pm$ 1.8 & 21.1 $\pm$ 28.2 & 133.0 $\pm$ 18.5
            & 5.5 $\pm$ 8.4 & 15.7 $\pm$ 33.0 & 8.2 $\pm$ 14.5
            & 0.8 $\pm$ 0.1 & 2.3 $\pm$ 0.1 & 4.4 $\pm$ 0.7
            & 11.4 $\pm$ 1.7 & 130.3 $\pm$ 38.5 & N/A \\
            % \cline{2-17}
            & \textit{PDDLStream}${+}$ & 0.5 $\pm$ 0.1 & 0.3 $\pm$ 0.0 & 0.2 $\pm$ 0.0 
            & 7.9 $\pm$ 1.8 & 4.2 $\pm$ 5.6 & 10.3 $\pm$ 5.3
            & 4.9 $\pm$ 5.9 & 1.1 $\pm$ 1.2 & 0.4 $\pm$ 0.2
            & 0.8 $\pm$ 0.1 & 0.7 $\pm$ 0.1 & 0.4 $\pm$ 0.0
            & 12.1 $\pm$ 1.6 & 3.1 $\pm$ 1.4 & 2.5 $\pm$ 0.4\\
            \hline
        \end{tabular}
    \end{subtable}

    \begin{subtable}{\linewidth}
        \caption{Robustness of Subgoal Sequence(s) from \textit{B3-\textit{Cooking}}}
        \label{subtable:kitchen_goal_generalize}
        \centering
        \begin{tabular}{|c|c|c|c|c|c|c|c|c|c|c|c|c|c|c|c|c|}
            \hline
            \multicolumn{2}{|c|}{} & \multicolumn{3}{c|}{\textbf{Object Geometry-Ingredient3}} & \multicolumn{3}{c|}{\textbf{Number of Objects-Ingredient3}} & \multicolumn{3}{c|}{\textbf{Number of Robots-Ingredient2}} & \multicolumn{3}{c|}{\textbf{Action Abstration-Ingredient2}} & \multicolumn{3}{c|}{\textbf{Action Set-Ingredient4}}\\
            \cline{3-17}
            \multicolumn{2}{|c|}{} & \textbf{Goal0} & \textbf{Goal1} & \textbf{Goal2} & \textbf{Goal0} & \textbf{Goal1} & \textbf{Goal2} & \textbf{Goal0} & \textbf{Goal1} & \textbf{Goal2} & \textbf{Goal0} & \textbf{Goal1} & \textbf{Goal2} & \textbf{Goal0} & \textbf{Goal1} & \textbf{Goal2} \\
            \hline
            \textbf{Success Rate} & \textit{PDDLStream} & 100 & 100 & 100  
            & 100 & 100 & 100
            & 70 & 30 & 30  
            & 100 & 100 & 100 
            & 70 & 0 & 0 \\
            % \cline{2-17}
            & \textit{PDDLStream}${+}$ & 100 & 100 & 100 & 100 & 100 & 100 & 100 & 100 & 100 & 100 & 100 & 100 & 100 & 100 & 100  \\
            \cline{1-2}
            \textbf{Time} [\textit{s}] & \textit{PDDLStream} & 0.9 $\pm$ 0.1 & 0.9 $\pm$ 0.1 & 0.9 $\pm$ 0.1 
            & 19.2 $\pm$ 8.0 & 25.8 $\pm$ 6.7 & 26.8 $\pm$ 5.0
            & 53.5 $\pm$ 41.2 & 55.3 $\pm$ 17.3 & 55.3 $\pm$ 17.3
            & 3.7 $\pm$ 0.1 & 3.7 $\pm$ 0.1 & 3.7 $\pm$ 0.1
            & 128.4 $\pm$ 12.5 & N/A & N/A \\
            % \cline{2-17}
            & \textit{PDDLStream}${+}$ & 0.4 $\pm$ 0.1 & 0.5 $\pm$ 0.1 & 0.4 $\pm$ 0.0 
            & 2.1 $\pm$ 1.2 & 1.8 $\pm$ 0.8 & 2.1 $\pm$ 0.5
            & 10.2 $\pm$ 7.9 & 17.1 $\pm$ 13.4 & 17.1 $\pm$ 13.4
            & 0.9 $\pm$ 0.2 & 0.9 $\pm$ 0.2 & 0.9 $\pm$ 0.2
            & 11.6 $\pm$ 2.2 & 8.5 $\pm$ 1.5 & 7.5 $\pm$ 4.0\\
            \hline
        \end{tabular}
    \end{subtable}
    \caption{Evaluation of the robustness of subgoal sequence(s) generated from past experiences \textit{Block B1} and \textit{Kitchen-B1} in five similar planning domains with different task variants. The 'N/A' symbol indicates tasks for which \textit{PDDLStream} fails to solve within 180-second timeout. }
    \label{tab:goal_generalization}
    \vspace{-0.5cm}
\end{table*}

\textbf{Robustness of Goal Decomposition}: As discussed in in Section~\ref{app:goal_decomposition}, we argue that the discovered subgoal sequences are a description of the implicit domain rules, expressed in the language of abstracted states. Therefore, they are robust to domain variants that do not reduce the set of domain rules previously considered. Below, we provide five specific domain variants that satisfy this condition and indicate when such robustness is beneficial. 

To quantify the benefits, we compare success rate and average planning time between \textit{PDDLStream} and \textit{PDDLStream} with goal decomposition (\textit{PDDLStream}${+}$). The comparison is conducted across domain variants that differ from the domains where the past experience for goal decomposition was gathered. Table~\ref{tab:goal_generalization} reports results for these five variants.

\subsubsection{V1-Object Geometry} In this variant, we changed the geometric sizes of blocks, ingredients, and blockers along the X or Y axis to generate different geometry changes. Correspondingly, some feasible grasp poses in the data-collection domain become infeasible in the current domain. However, changes in motion-planning feasibility do not alter the fundamental domain rules, so the original subgoal sequences remain applicable in similar planning domains. As shown in Tables~\ref{subtable:block_goal_generalize} and~\ref{subtable:kitchen_goal_generalize}, we successfully transfer the discovered subgoal sequences to all evaluated tasks. Specifically, \textit{PDDLStream}${+}$ achieved consistent success on the variant tasks using the original subgoal sequences and attained a significantly shorter average planning time than \textit{PDDLStream}. Therefore, we can use the same subgoal sequences to accelerate multi-object manipulation tasks with changing object geometry.

\subsubsection{V2-Number of Objects} In this variant, we varied the number of objects by introducing six additional blocks for tasks in the \textit{Tower Construction} domain and three additional ingredients in the \textit{Cooking} domain. The introduction of additional objects may change the feasibility of many motion planning subproblems and therefore requires task-and-motion replanning, even though the task goals remain the same. However, since the task goals and domain rules do not change with the number of objects, the original subgoal sequences from the simpler domain remain valid in this more complex scenario. As shown in Table~\ref{tab:goal_generalization}, the original subgoal sequences consistently solved the new tasks with generally shorter planning times, accelerating multi-object manipulation planning with online changes in the number of objects.

\subsubsection{V3-Number of Robots} In this variant, we change the number of robots in the planning domain by extending the original single-robot domain to a dual-arm planning domain. With the addition of a second robot, the potential interactions between robots and objects increase, changing the feasibility of motion planning subproblems and often increasing the overall planning time. However, adding robots to the planning domain does not alter the domain rules or the original subgoal sequences. Consequently, \textit{PDDLStream}${+}$ consistently solved all planning tasks with significantly shorter planning times, accelerating multi-object manipulation tasks with varying numbers of robots.

\subsubsection{V4-Action Abstraction}
In this variant, we evaluate the robustness of subgoal sequences for planning domains where actions are abstracted at different hierarchical levels compared to the original planning domain. Specifically, we replace the original action set $\mathcal{A}={\textit{pick},\ \textit{place},\ \textit{unstack},\ \textit{stack}}$ in \textit{Tower Construction} with an alternative set $\mathcal{A}_{1}={\textit{pick-place},\ \textit{pick-stack},\ \textit{unstack-place}}$. In \textit{Cooking}, we similarly replace \textit{wash} and \textit{cut} with a single abstracted action, \textit{prepare}. These new actions are obtained by merging certain original actions, making some subgoals unreachable because they correspond to intermediate states within the newly abstracted actions. Nevertheless, at any level of action abstraction, the space of feasible state trajectories remains the same, since the domain rules are not affected by the abstraction hierarchy. Consequently, the original subgoal sequences generalize effectively to domains with actions defined at higher levels of abstraction. To evaluate this, we make a slight modification to \textit{PDDLStream}${+}$: when the current subgoal is infeasible due to action abstraction, the system selects the next temporally closest consecutive subgoal. 

From the experimental results, we observe consistent success in solving the original tasks with different action abstractions, and the planning time is significantly shorter than with standard \textit{PDDLStream}. This demonstrates the robustness of goal decomposition across variations in the action-abstraction hierarchy, enabling reuse of learned subgoal sequences in planning domains with different action abstractions.

\subsubsection{V5-Action Set} An intelligent robot assistant learns new skills in a lifelong manner and therefore has an incrementally growing action set. Accordingly, our learned heuristics should be robust to changes in the action set. To evaluate robustness to this domain variant, we compare the performance of \textit{PDDLStream} and \textit{PDDLStream}${+}$ in tasks where new objects of unseen types and corresponding new actions are introduced. We introduce \textit{tool} as a new object type and \textit{push} as a new action to \textit{Tower Construction}. We also add \textit{lid} as a new object type and \textit{lid-place} as a new action to \textit{Cooking}. As shown in Table~\ref{tab:goal_generalization}, \textit{PDDLStream}${+}$ consistently solves all problems that \textit{PDDLStream} alone struggles to complete within the 180,s timeout. Importantly, \textit{PDDLStream}${+}$ requires significantly less planning time and achieves a higher success rate. These results demonstrate the robustness of goal decomposition to increased action and object sets, enabling the solution of multi-object manipulation tasks with varied skill sets for handling unseen objects.

With experiment results above, we conclude that: with our goal decomposition method, the generated subgoal sequences show robustness to various domain variants, enabling multi-object manipulation planning with changing object geometry, object number, robot number, and action abstraction.

\textbf{Robustness of Computational Distance}: In this section, we evaluated the effectiveness and robustness of the learned computational distance function to predict the temporally closest subgoal from a random disturbed state. Since the computational distance function is intended for use with the subgoal sequences, we therefore demonstrate that our computational distance function exhibits similar robustness to domain variations as our goal decomposition method. Consequently, we designed our numerical experiments with the same domain variants as in Table \ref{tab:goal_generalization} for consistency. Additionally, as \textit{Goal 0} in the block domains is not decomposable (thus not requiring subgoal prediction), we thus simplified our analysis by focusing on \textit{Goal 1} and \textit{Goal 2} in \textit{Tower Construction} and \textit{Cooking} domains.

\begin{figure*}[thb!]
    \centering
    \begin{subfigure}{0.45\linewidth}  % Adjusted width for better visibility
        \centering
        \includegraphics[width=\linewidth]{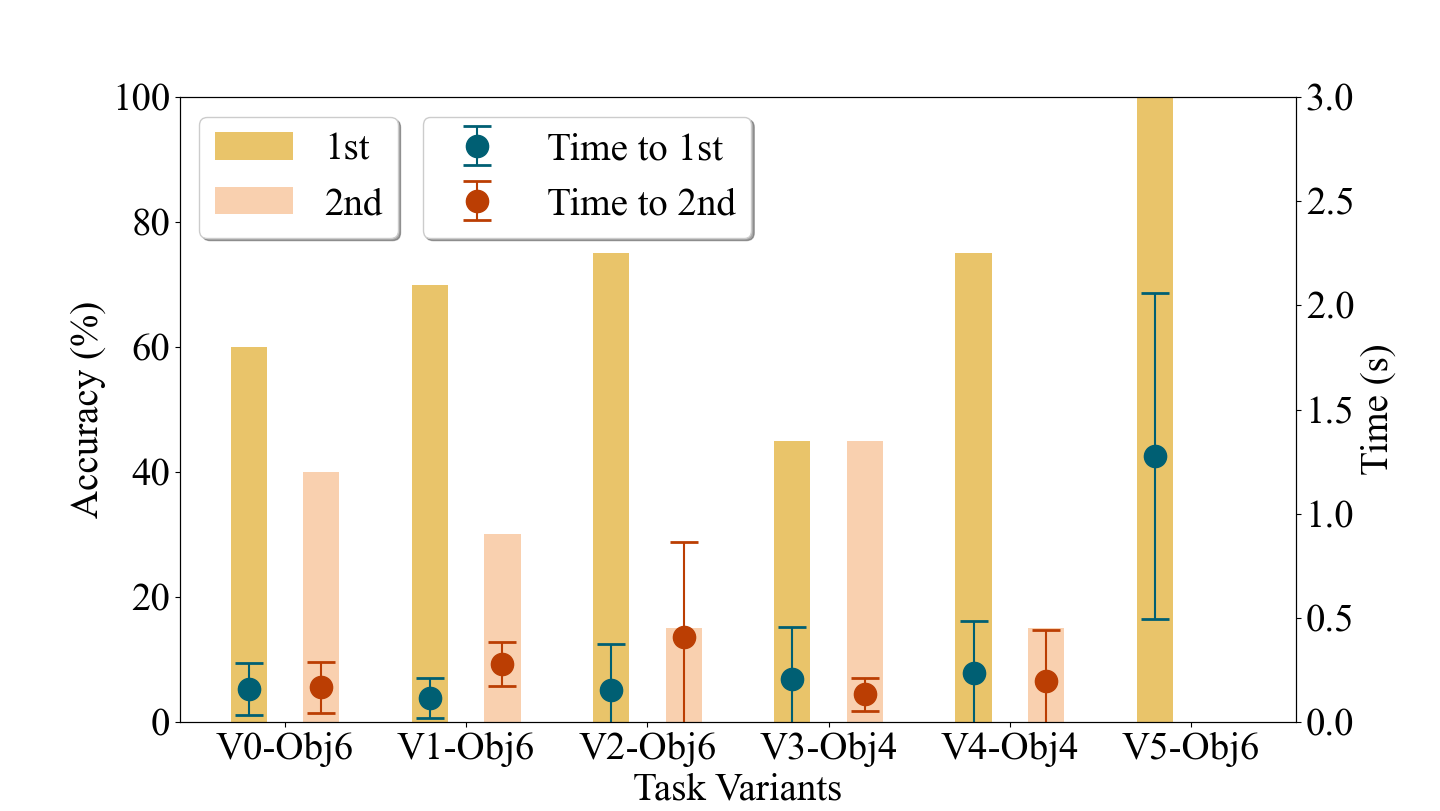}
        \caption{Tower Construction under Task Goal 1}
        \label{fig:block_temporal_dist_goal1}
    \end{subfigure}%
    \hspace{0.02\linewidth}
    \begin{subfigure}{0.45\linewidth}
        \centering
        \includegraphics[width=\linewidth]{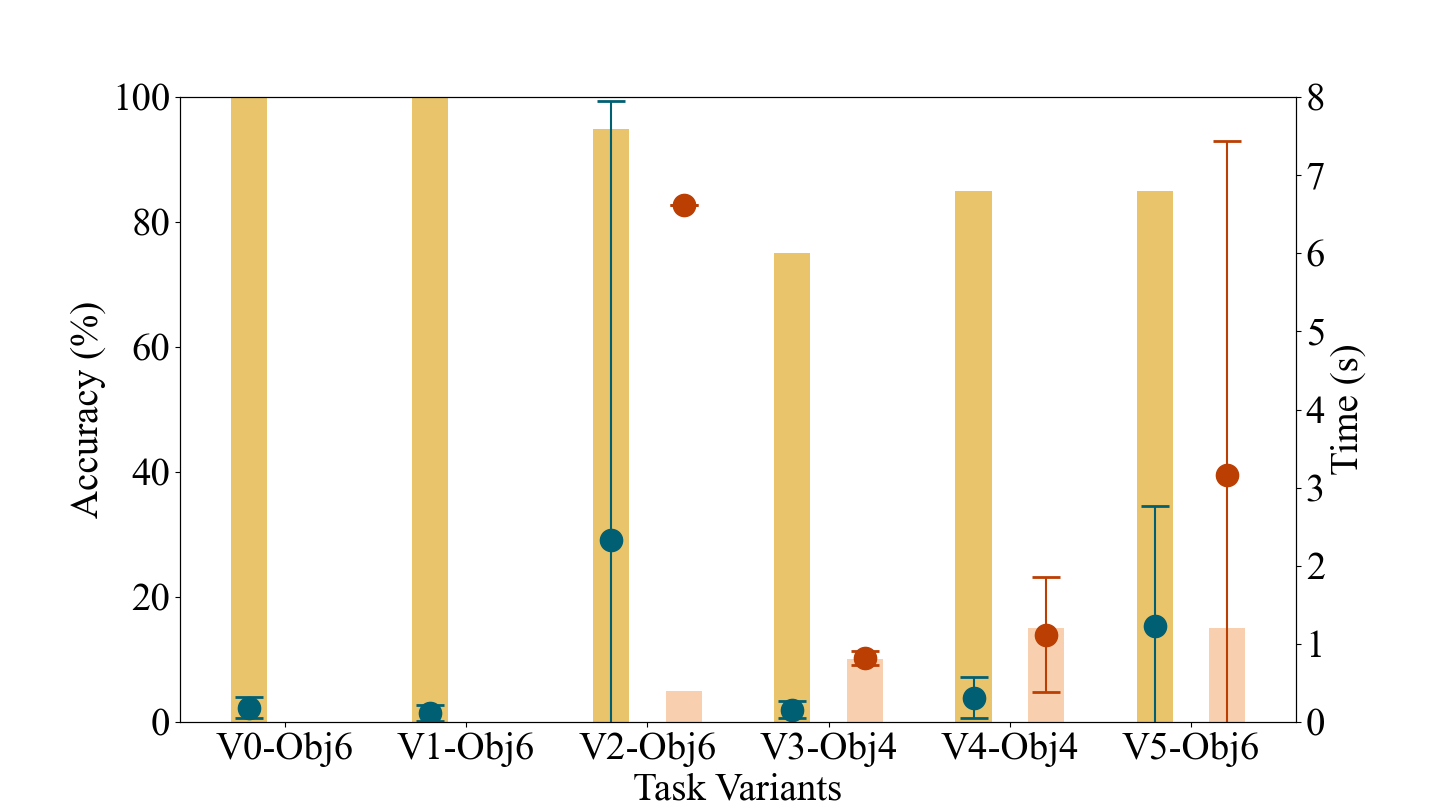}
        \caption{Tower Construction under Task Goal 2}
        \label{fig:block_temporal_dist_goal2}
    \end{subfigure}%
    \hspace{0.02\linewidth}
    \begin{subfigure}{0.45\linewidth}  % Adjusted width for better visibility
        \centering
        \includegraphics[width=\linewidth]{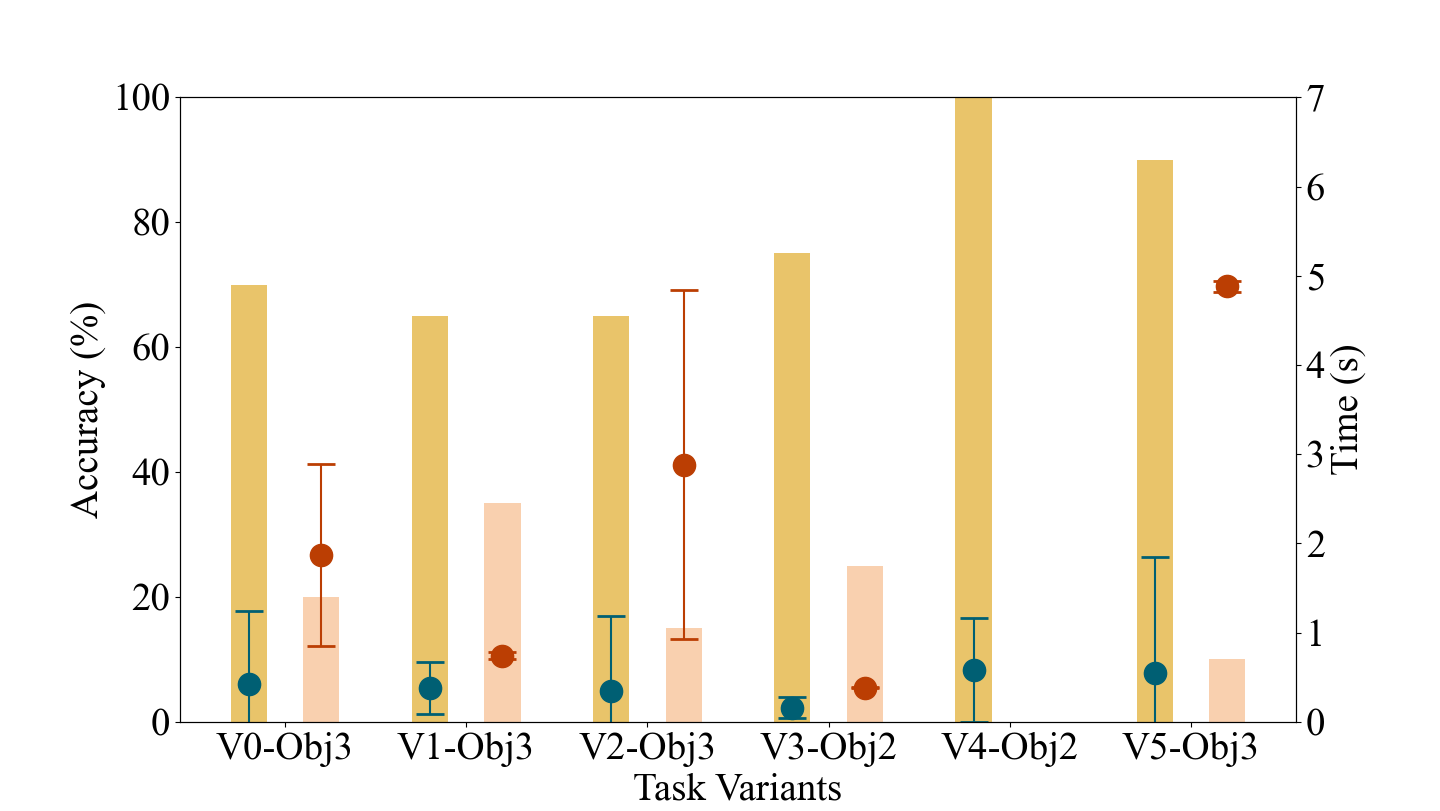}
        \caption{Cooking under Task Goal 1}
        \label{fig:kitchen_temporal_dist_goal1}
    \end{subfigure}%
    \hspace{0.02\linewidth}
    \begin{subfigure}{0.45\linewidth}
        \centering
        \includegraphics[width=\linewidth]{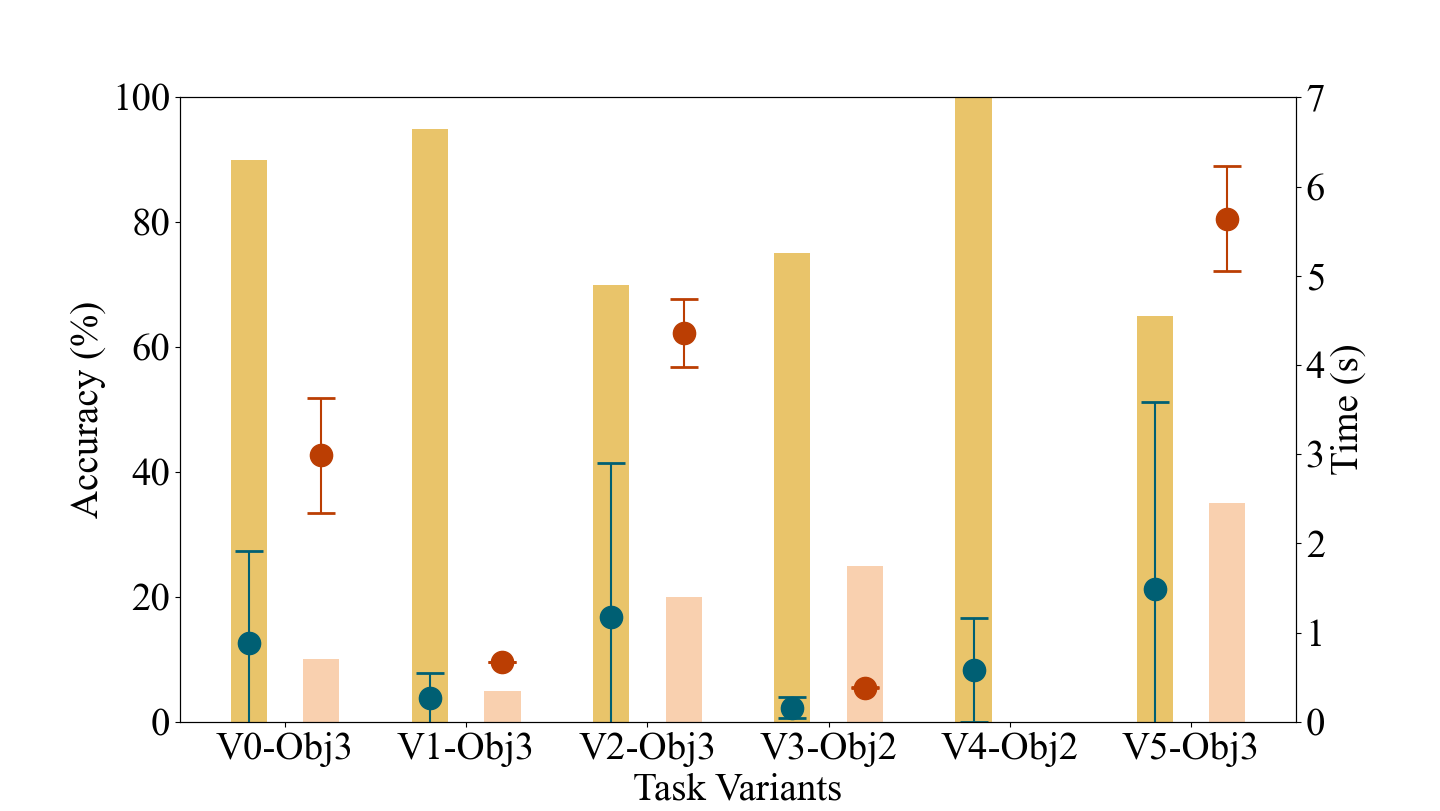}
        \caption{Cooking under Task Goal 2}
        \label{fig:kitchen_temporal_dist_goal2}
    \end{subfigure}%
    \hspace{0.02\linewidth}
    \caption{Evaluation of the robustness of learned computational distance function. All four figures share the same legends. Bar charts with legend \textit{1st} and \textit{2nd} show the accuracy of finding the 1st and 2nd temporal closest subgoal. Error bars with legend \textit{Time to 1st/2nd} show the actual planning time for reaching the specific subgoal.
    Experiment setup is same to the one for Table \ref{tab:goal_generalization}. \textit{V0} corresponds to domains with past experience from which the computational distance is trained: \textit{Block B1} and \textit{Kitchen B1} for Block and Kitchen domains. From \textit{V1} to \textit{V5}, they refers to domain variants with object geometry, number of objects, number of robots, action abstraction, and action set, respectively. Computational distance from \textit{V0} effectively find the temporally closest subgoals in both \textit{V0} and other domains, showcasing its effectiveness and robustness. 
    }
    \vspace{-0.5cm}
    \label{fig:effect_temporal_dist}
\end{figure*}

Note that the planning time of \textit{PDDLStream} is not deterministic, as we observe that the distribution of planning times for consecutive subgoals may overlap. This overlap can result in the temporally closest subgoal occasionally appearing as the second closest due to planning-time variation. Therefore, we use the average accuracy in selecting the \textit{first two} temporally closest subgoals to assess the temporal distance function’s effectiveness across various planning domains. The robustness is then evaluated by testing the temporal distance function trained on \textit{B1-Tower Construction} and \textit{B3-Cooking} domains in other domain variants, with variations in object geometry (V1), number of objects (V2), number of robots (V3), action abstraction (V4), and action sets (V5).

The robustness of the computational distance function arises primarily from three factors: 1) the generalization capability of the GNN model, 2) the \textit{relative} distance definition, and 3) subgoals with \textit{sequential orders}. For a task involves six blocks in the \textit{Tower construction} domain, possible initial states far exceed the number of past experiences (40). However, the computational distance function generalizes to other states, as these states may share symmetrical graph properties with those in the dataset, enabling accurate predictions of distance between two states (exploiting the generalization capability of the GNN model). Furthermore, if an object is misclassified as essential or non-essential for one subgoal $\bm{G}_{i}$, this misclassification will likely persist in subsequent subgoals $\bm{G}_{i+z}$ ($z \in [1, Z-i]$), as the sequential structure of subgoals requires passing through $\bm{G}_{i}$ before reaching $\bm{G}_{i+z}$. Thus, capturing \textit{relative} temporal distance reduces bias from such misclassifications when subgoals have specific \textit{sequential orders}. 

\begin{figure*}[thb!]
    \centering
    \begin{subfigure}{0.45\linewidth}  % Adjusted width for better visibility
        \centering
        \includegraphics[width=\linewidth]{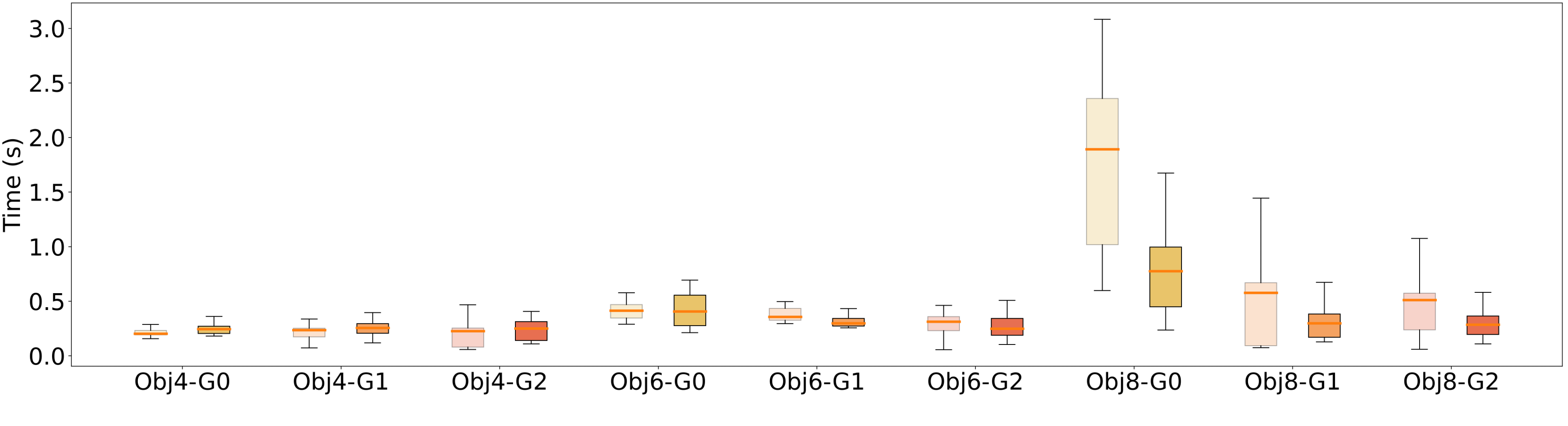}
        \caption{Block B1-Planning Time}
        \label{fig:block_obj_reuduct_time}
    \end{subfigure}%
    \hspace{0.02\linewidth}
    \begin{subfigure}{0.45\linewidth}
        \centering
        \includegraphics[width=\linewidth]{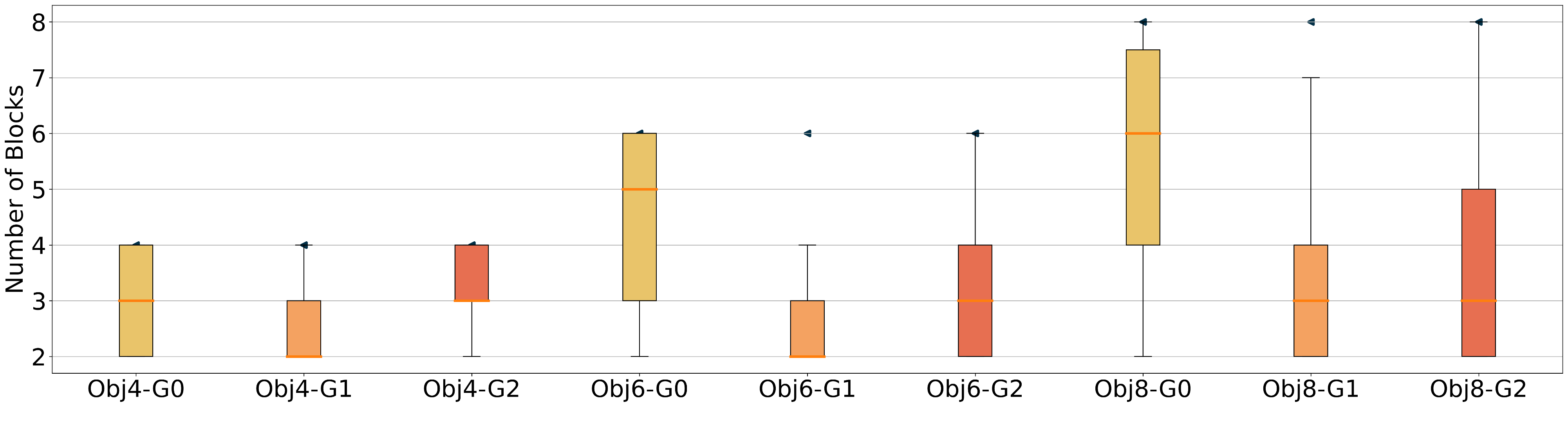}
        \caption{Block B1-Object Count}
        \label{fig:block_obj_reduct_obj}
    \end{subfigure}%
    \hspace{0.02\linewidth}
    \begin{subfigure}{0.45\linewidth}  % Adjusted width for better visibility
        \centering
        \includegraphics[width=\linewidth]{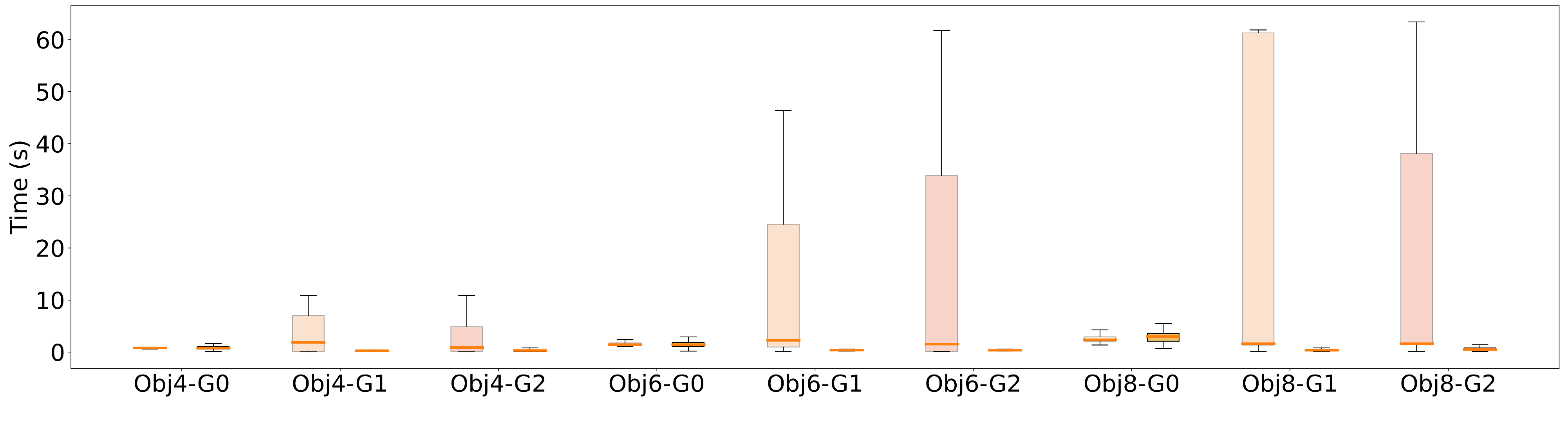}
        \caption{Block B1 with Additional Blocks-Planning Time}
        \label{fig:block_v2_obj_reduction_time}
    \end{subfigure}%
    \hspace{0.02\linewidth}
    \begin{subfigure}{0.45\linewidth}
        \centering
        \includegraphics[width=\linewidth]{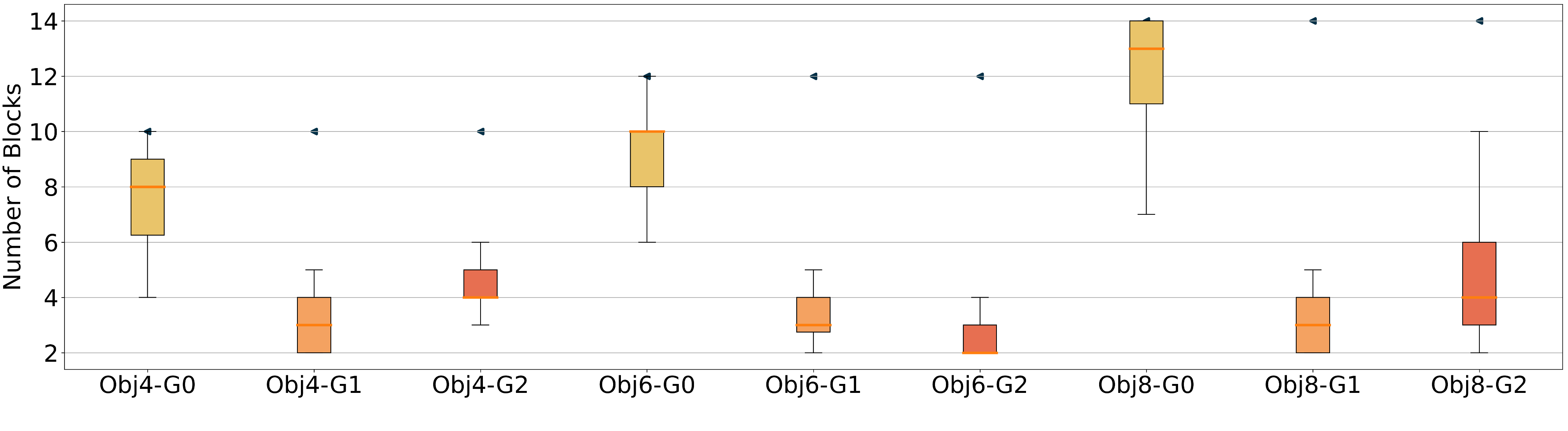}
        \caption{Block B1 with Additional Blocks-Object Count}
        \label{fig:block_v2_obj_reduction_obj}
    \end{subfigure}%
    \hspace{0.02\linewidth}
    \begin{subfigure}{0.45\linewidth}  % Adjusted width for better visibility
        \centering
        \includegraphics[width=\linewidth]{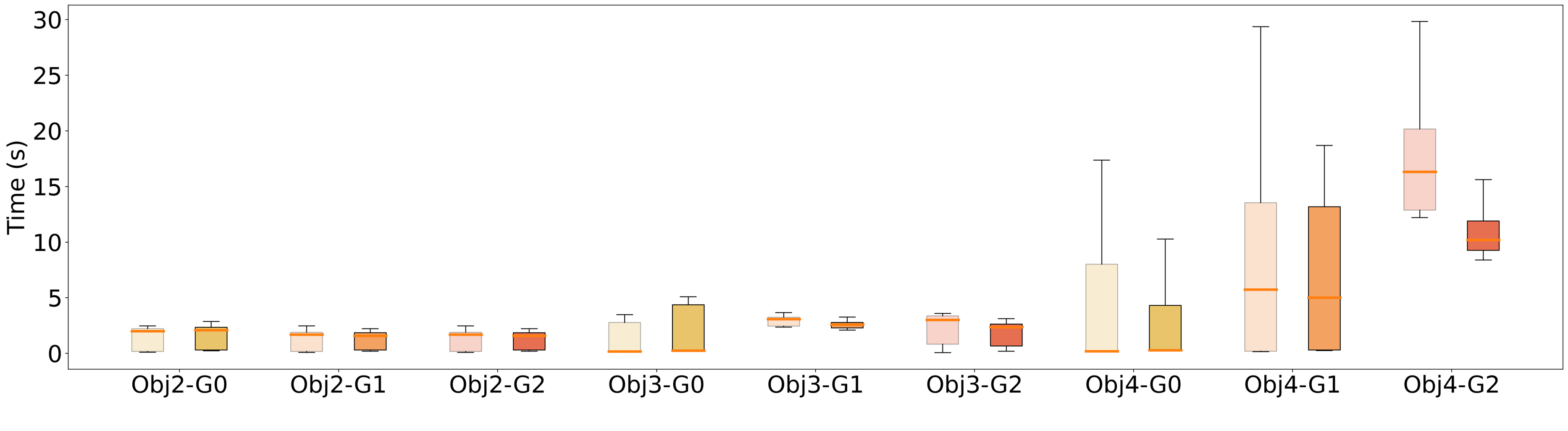}
        \caption{Kitchen B1-Planning Time}
        \label{fig:kitchen_obj_reduction_time}
    \end{subfigure}%
    \hspace{0.02\linewidth}
    \begin{subfigure}{0.45\linewidth}
        \centering
        \includegraphics[width=\linewidth]{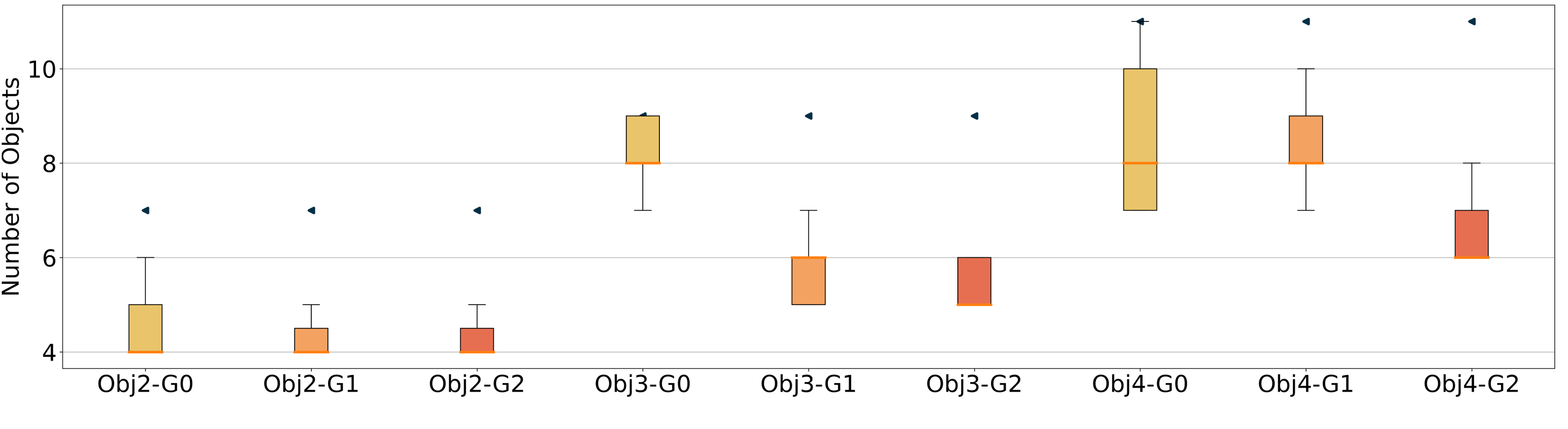}
        \caption{Kitchen B1-Object Count}
        \label{fig:kitchen_obje_reduction_onj}
    \end{subfigure}%
    \hspace{0.02\linewidth}
    \begin{subfigure}{0.45\linewidth}  % Adjusted width for better visibility
        \centering
        \includegraphics[width=\linewidth]{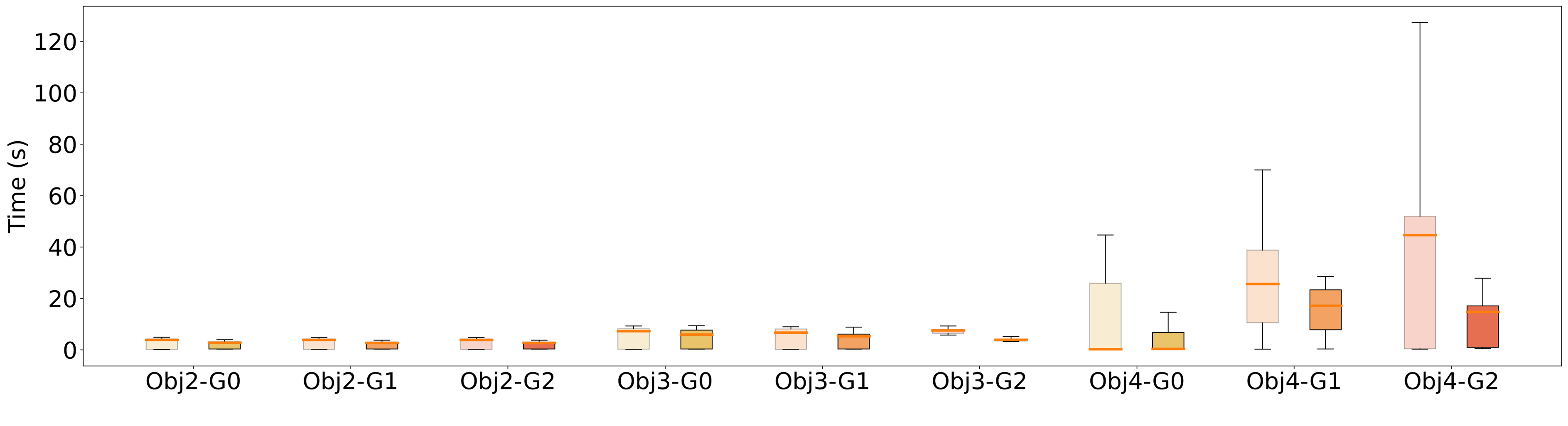}
        \caption{Kitchen B1 with Additional Blockers-Planning Time}
        \label{fig:kitchen_v2_object_reduction_time}
    \end{subfigure}%
    \hspace{0.02\linewidth}
    \begin{subfigure}{0.45\linewidth}
        \centering
        \includegraphics[width=\linewidth]{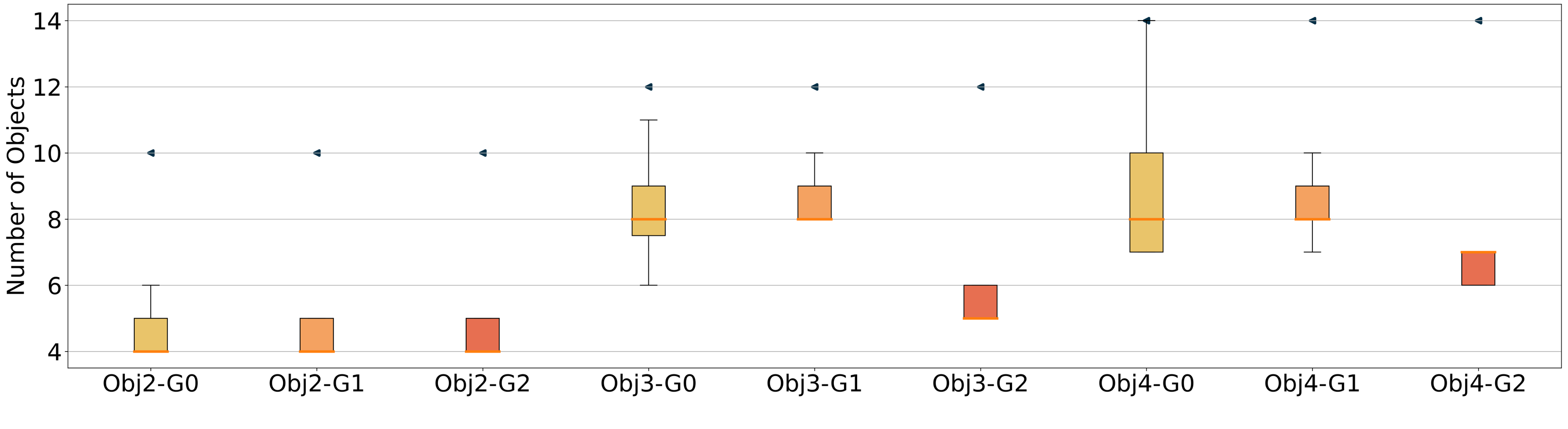}
        \caption{Kitchen B1 with Additional Blockers-Object Count}
        \label{fig:kitchen_v2_object_reduction_obj}
    \end{subfigure}%
     \caption{Evaluation of the effectiveness and robustness of the object reduction strategy. Left figures compare the planning time of \textit{PDDLStream} with and without object reduction. Right figures show the corresponding distributions of essential objects when using object reduction, with black triangles indicating the constant numbers of blocks without object reduction. In each left figures, we illustrate the planning time of with object reduction at the right side of each X labels with darker colors then those of without object reduction. A label with name \textit{Obj4-G0} refers to planning with 4 cubes with characters for reaching the closest subgoal of task goal 0.}
    \label{fig:effect_obj_reduction}
\end{figure*}
As shown in Figure~\ref{fig:effect_temporal_dist}, the distance function learned in \textit{V0} reliably identifies the two temporally closest subgoals in both \textit{V0} and the other variants, demonstrating its effectiveness. In terms of robustness:
\begin{itemize}
    \item \textit{V1-Object Geometry}: Changing object geometry renders some previously feasible motion-planning subproblems infeasible. However, this increased difficulty is reflected in the \textit{absolute} distances to affected subgoals. Using the \textit{relative} distance, the method still identifies the closest subgoal correctly. Consequently, the learned distance achieves 100\% top-2 accuracy across different task-goal settings.
    \item \textit{V2-Object Number}: When the number of objects increases, the input graph to the trained GNN grows. The GNN’s generalization capacity supports accurate predictions as objects are added, consistent with prior findings~\cite{silver2021planning, khodeir2023learning}. Correspondingly, Figure~\ref{fig:effect_temporal_dist} shows that the learned distance function accurately predicts the two closest subgoals from randomly disturbed states for tasks under variant \textit{V2}.
    \item \textit{Others}: Our distance definition approximates the planning complexity of transferring disturbed states to subgoals and is, by design, largely independent of the number of robots (V3) and the abstraction level or extension of the action set (V4, V5). While these variants influence absolute planning time for reaching subgoals, the relative distance combined with a fixed subgoal order remains sufficient for selecting the closest subgoals, as evidenced in Figure~\ref{fig:effect_temporal_dist}.
\end{itemize}

However, the robustness of our computational distance relies on the sequential order of subgoals, which limits accuracy for task goals that admit multiple valid subgoal sequences. Accordingly, for goals with multiple sequences (e.g., \textit{Goal~1}), we observe a higher probability of selecting the second-closest subgoal as the target closest one, compared with goals that have a single sequence (e.g., \textit{Goal~2}). As future work, it would be beneficial to develop improved metrics for approximating computational distance in settings with branching subgoal structures.

\textbf{Robustness of Object Reduction}:
The proposed object reduction function aims to filter out non-essential environmental objects while reaching to the predicted temporally closest subgoal. Therefore, we need to predict the object importance as accurately as possible. Although we use the same trained GNN model as the temporal distance, the \textit{relative} distance definition and sequential order of subgoals do not contribute to the robustness for object reduction. Consequently, the robustness of object reduction relies on the generalization capability of GNN model. However, our parallel planning with increased object sets ensures that all the environment objects are considered in the worst case. Therefore, no matter how the domain varies, the feasibility of the problem is not changes under object reduction. For domain variants where GNN can be generalized to, the predicted object importance is accurate enough, we will gain more planning efficiency improvement. For variants, it can be generalize to, we observe the same planning efficiency as without object reduction.

We thus focus on evaluating the effectiveness of object reduction in terms of filtering out non-essential objects in the \textit{V0} domains and its robustness to increasing numbers of objects (with seen objet types) in the \textit{V2} domains. We compared the average planning time for reaching the predicted not-yet-reached closest subgoal and the corresponding number of essential objects when with and without the proposed object reduction strategy, across 100 random initial states in various task goals and object counts.

Our experiments demonstrate that the object reduction strategy consistently preserves the feasibility of subgoal-reaching tasks, as evidenced by the successful completion of all subgoal-reaching problems when combined with parallel subproblem planning. Moreover, as shown in the object count figures in Figure \ref{fig:effect_obj_reduction}, the object reduction strategy effectively reduces the number of objects involved in planning, even for non-decomposable subgoals such as \textit{Goal 0} in the \textit{Block-B1} domain. Correspondingly, the planning time required by \textit{PDDLStream} is substantially reduced, especially in environments with a high initial number of objects (e.g., \textit{B1-Tower Construction} with eight cubes and \textit{B3-Cooking} with four ingredients). The impact becomes more pronounced when additional objects are introduced into these domains.

\end{document}